\documentclass{article}
\usepackage{iclr2027_conference,times}

\usepackage[utf8]{inputenc}
\usepackage[T1]{fontenc}
\usepackage{hyperref}
\usepackage{url}
\usepackage{booktabs}
\usepackage{amsfonts}
\usepackage{amsmath}
\usepackage{nicefrac}
\usepackage{microtype}
\usepackage{xcolor}
\usepackage{graphicx}
\usepackage{multirow}
\usepackage{wrapfig}
\usepackage{amssymb}
\usepackage{pifont}
\usepackage{color,colortbl}
\usepackage{CJKutf8}
\usepackage{xspace}

\newcommand{\cmark}{\ding{51}}
\newcommand{\xmark}{\ding{55}}

\newcommand{\benchname}{DramaVQA\xspace}
\newcommand{\dramas}{12\xspace}
\newcommand{\QAs}{20K\xspace}
\newcommand{\fullQAs}{20,000\xspace}

\newcommand{\tmedium}{$\mathtt{Medium}$\xspace}
\newcommand{\tlong}{$\mathtt{Long}$\xspace}
\newcommand{\textralong}{$\mathtt{Extra\text{-}Long}$\xspace}
\newcommand{\tultralong}{$\mathtt{Ultra\text{-}Long}$\xspace}
\newcommand{\qtvp}{\textsf{visual perception}\xspace}
\newcommand{\qtsu}{\textsf{spatial understanding}\xspace}
\newcommand{\qtai}{\textsf{action \& interaction}\xspace}
\newcommand{\qteu}{\textsf{event understanding}\xspace}
\newcommand{\qttu}{\textsf{temporal understanding}\xspace}
\newcommand{\qtrr}{\textsf{relation \& reasoning}\xspace}
\newcommand{\qtco}{\textsf{counting}\xspace}

\newcommand{\promptsection}[1]{\vspace{0.5em}\noindent\textbf{#1:}\par\nobreak\vspace{0.25em}}
\newcommand{\modelname}{DramaVQA\xspace}
\definecolor{lightgray}{gray}{0.85}

\title{\benchname: Probing Detailed Understanding of Ultra-Long TV Dramas}

\newcommand{\dramavqaauthorblock}{\parbox[t]{\dimexpr\textwidth-2\tabcolsep\relax}{%
\centering\normalfont
\setlength{\parindent}{0pt}%
\setlength{\parskip}{0pt}%
{\fontsize{11}{14}\selectfont\bfseries
Xinyue~Huo, Pengfei~Chen, Jiannan~Ge, Kaiwen~Duan, Jiacheng~Shao, Yunning~Cao, Lingxi~Xie$^{*,\dagger}$, Qi~Tian$^{\dagger}$\par}
\vspace{3pt}
{\fontsize{10}{12}\selectfont Huawei Technologies Ltd\par}
\vspace{3pt}
{\fontsize{9}{11}\selectfont
\{huoxinyue2, chenpengfei64, gejiannan2, tian.qi1\}@huawei.com\par
kaiwenduan@outlook.com, shaojiacheng2@h-partners.com, ytcaoyunning@gmail.com, 198808xc@gmail.com\par}}}
\author{\dramavqaauthorblock}

\iclrfinalcopy 

\usepackage{booktabs,tabularx,array,multirow,pifont,xcolor,colortbl}

\providecommand{\benchname}{Benchmark Name}

\begin{document}

\maketitle
\lhead{} 

\begingroup
\renewcommand{\thefootnote}{\fnsymbol{footnote}}
\begin{NoHyper}
\footnotetext[1]{Project leader: Lingxi Xie.}
\footnotetext[2]{Corresponding authors: Lingxi Xie and Qi Tian.}
\end{NoHyper}
\endgroup

\begin{abstract}
Despite recent progress in long-form video understanding, current benchmarks mostly focus on short clips or hour-scale videos. This leaves a major gap in evaluating ultra-long video understanding, where videos often last tens of hours. To bridge this gap, we introduce \benchname, a large-scale benchmark built upon long-form TV dramas. Containing approximately \QAs question-answering (QA) pairs, the scale of \benchname exceeds previous long-video datasets, such as Video-MME and LVBench, by an order of magnitude. Our benchmark has two main features that make it stand out from existing datasets. First, we establish a scalable, LLM-assisted annotation pipeline that uses multimodal large language models (MLLMs) to generate complex questions, followed by strict human checking to ensure high data quality. Second, \benchname introduces a multi-level time structure; queries are designed across different video lengths, with the hardest questions requiring overall understanding across the entire drama. Extensive evaluations show that state-of-the-art MLLMs face major challenges when tested on these very long videos, highlighting the need for agentic AI algorithms capable of managing long-term memory. \textit{The project is available at \url{https://github.com/xinyuehuo/DramaVQA}.}

\end{abstract}

\section{Introduction}
\label{introduction}

The rapid evolution of multimodal large language models (MLLMs) has greatly expanded the limits of video understanding. Recent state-of-the-art models, such as Gemini-3-Pro~\cite{team2023gemini} and the Qwen-VL series~\cite{bai2025qwen2,bai2025qwen3}, now support very large context windows (\textit{e.g.}, 128K input tokens). This allows them to achieve remarkable accuracy on question-answering (QA) tasks for hour-long videos previously considered impossible. However, as model capacities grow, the community lacks a benchmark that pushes the frontier toward ultra-long contexts, specifically videos lasting tens of hours, where processing the entire video at once is no longer computationally practical.

To address this gap, we introduce \benchname, a large-scale benchmark built upon long-form TV dramas to encourage research in ultra-long video understanding. TV dramas serve as an ideal testbed because they feature continuous, detailed storylines driven by character growth and long-range cause-and-effect relationships. This structure allows us to create challenging questions that require finding key information across dozens of episodes. As a result, it rigorously tests MLLM capabilities across multiple dimensions, including visual recognition, temporal grounding, and high-level summarization.

The main hurdle in creating such a benchmark is the high cost of human annotation and the difficulty of ensuring quality control. To overcome this, we propose a scalable, LLM-assisted annotation pipeline with four key stages: (1) extracting structured metadata from the dramas (\textit{e.g.}, subtitles, facial identities, speaker identities) and using MLLMs to generate multi-level video captions; (2) using LLMs to generate high-quality candidate QA pairs from these captions; (3) employing human annotators to carefully filter and correct wrong or ambiguous QAs; and (4) using LLMs one last time to verify that the remaining questions are unique across the entire drama. This method allows \benchname to reach an unprecedented scale of approximately \QAs QA pairs, an order of magnitude larger than existing long-video datasets like Video-MME~\cite{fu2025video} and LVBench~\cite{wang2025lvbench}. Table~\ref{tab:benchmarks} shows a detailed comparison between \benchname and recent benchmarks.

This construction process naturally gives \benchname a multi-level time structure. Each QA pair can be tested using different input video lengths, categorized into four distinct temporal tracks: \tmedium (averaging 7.6 minutes), \tlong (averaging 43.1 minutes), \textralong (averaging 9.3 hours), and \tultralong (the full drama, averaging 38.9 hours). Notably, the latter two tracks are much longer than existing benchmarks (\textit{e.g.}, the average durations of Video-MME-Long and LVBench are only 0.69 and 1.14 hours, respectively). Our extensive evaluation of state-of-the-art MLLMs shows a significant performance drop as the video length increases. While current models perform reasonably well on the \tmedium track, their accuracy drops significantly on the \tlong and track, and falls to near-random guessing levels on the \textralong and \tultralong tracks. This sharp drop shows a major weakness in current architectures: they rely too heavily on brute-force context scaling and cannot handle ultra-long videos.

\begin{table}[!t]
\centering
\resizebox{0.98\linewidth}{!}{
\begin{tabular}{lcccccc}
\hline
\textbf{Benchmark} & \textbf{\#QAs} [per video] & \textbf{Avg Dura.} & \textbf{MG} & \textbf{VM} & \textbf{TG} & \textbf{Annotation} \\
\hline
TVQA-test~\cite{lei2018tvqa} & 15,253 [1.0] & 0.003h & \xmark & \xmark & -- & Manual \\
ActivityNet-QA~\cite{yu2019activitynet} & 800 [1.0] & 0.03h & \xmark & \xmark & -- & Manual \\
MVBench~\cite{li2024mvbench} & 4,000 [1.1] & 0.04h & \xmark & \xmark & -- & Auto \\
EgoSchema~\cite{mangalam2023egoschema} & 5,063 [1.0] & 0.05h & \xmark & \xmark & -- & Auto+Manual \\
MovieQA~\cite{tapaswi2016movieqa} & 6,462 [1.0] & 0.06h & \xmark & \xmark & -- & Manual \\
LongVideoBench~\cite{wu2024longvideobench} & 6,678 [1.8] & 0.13h & \xmark & \cmark & -- & Manual \\
MovieChat-1K~\cite{song2024moviechat} & 13,000 [13.0] & 0.16h & \xmark & \xmark & -- & Manual \\
MoVQA~\cite{zhang2025lvbench} & 20,061 [200.6] & 0.28h & \cmark & \xmark & \xmark & Manual \\
CG-Bench~\cite{chen2024cg} & 12,129 [10.0] & 0.45h & \cmark & \cmark & \cmark & Manual \\
Video-MME-L~\cite{fu2025video} & 900 [3.0] & 0.69h & \xmark & \cmark & \xmark & Manual \\
LVBench~\cite{wang2025lvbench} & 1,549 [15.0] & 1.14h & \xmark & \xmark & \xmark & Manual \\
\hline
\textbf{\benchname}-UL & \textbf{\fullQAs} [\textbf{1,666.7}] & \textbf{38.85h} & \cmark & \cmark & \cmark & Auto+Manual \\
\hline
\end{tabular}
}
\caption{\textbf{\benchname} excels in \textbf{QA number}, \textbf{average duration}, and other attributes among recent long-form video benchmarks. \textbf{MG}: multi-granular (each QA supports different durations); \textbf{VM}: video metadata available (subtitles, faces, speaker data, hierarchical video captions); \textbf{TG}: temporal grounding provided (only for 0.2h+ videos).}
\label{tab:benchmarks}
\end{table}

Beyond providing a strict testbed for zero-shot video QA, \benchname pushes the field forward in another important way. 
First, we will release over 400K high-quality QA pairs. Although many were filtered from the final benchmark for failing the strict uniqueness test, all these QAs have been human-validated and include precise timestamps, offering a highly valuable, large-scale dataset for MLLMs. 
Unlike traditional benchmarks that provide only 3--20 questions per video, \benchname supplies an average of 1,600+ questions per drama. This high density of queries over massive time scales challenges the community to move beyond simple, single-pass processing. Instead, it motivates a necessary shift toward memory-efficient long video understanding, where models must organize video streams into compact, easily searchable memory formats to perform complex reasoning within practical computational limits.

\section{Related Work}
\label{related_work}

\noindent \textbf{Multimodal large language models} (MLLMs).
The field of computer vision has greatly changed with the rise of MLLMs~\cite{hurst2024gpt,chen2024internvl,wang2024qwen2,bai2025qwen2,bai2025qwen3}. By building on the strong abilities of basic LLMs~\cite{touvron2023llama,achiam2023gpt,team2023gemini,guo2025deepseek,yang2025qwen3}, these models convert visual signals into discrete tokens, easily combining them with text prompts~\cite{liu2023visual,zhu2023minigpt,kim2024openvla,tong2024cambrian}. This step has moved the field past simple, closed-set tasks (\textit{e.g.}, basic detection or image classification) into open-ended video question answering~\cite{maaz2024video,lin2024video,zhang2024llava,li2024llavanextinterleave,li2024llavaonevision}. Furthermore, ongoing model improvements allow them to process much larger sequences of visual tokens, opening up new ways to understand longer videos.

\noindent \textbf{Benchmarks for long-form video understanding.}
Since current models have largely mastered static images and short video clips, research has shifted toward longer video analysis. Recent large-scale datasets focused on hour-long videos and complex QA tasks (\textit{e.g.}, EgoSchema~\cite{mangalam2023egoschema}, LongVideoBench~\cite{wu2024longvideobench}, Video-MME~\cite{fu2025video}, LVBench~\cite{wang2025lvbench}, CG-Bench~\cite{chen2024cg}) have created major challenges for modern vision systems. The main trend among these datasets is the steady increase of the input length (as shown in Table~\ref{tab:benchmarks}), which challenges MLLMs to integrate massive amounts of video information into their context windows. Meanwhile, some benchmarks~\cite{cheng2025video,liu2025videoreasonbench} use minute-level videos paired with complex logical questions to test the reasoning limits of these models. Although LongTimeScope~\cite{zohar2025apollo2} and Ego-R1 Bench~\cite{tian2025ego} cover much longer videos, they contain only 450 and 300 evaluation questions, respectively, with limited coverage of fine-grained understanding within individual videos.  To address this, our research delivers the largest benchmark built to date for rigorous, ultra-long video reasoning, constructed specifically from long-form TV dramas. While datasets like MovieQA~\cite{tapaswi2016movieqa}, MovieChat~\cite{song2024moviechat}, DramaQA~\cite{choi2021dramaqa}, and MoVQA~\cite{zhang2025lvbench} also focus on video understanding using movie data, their video durations are still at the sub-hour level. By using multi-episode TV dramas, we make it possible to extend the video duration to the 10-hour level.

\noindent \textbf{Advanced solutions for long-form video understanding.}
To mitigate inherent context limitations of MLLMs, existing methods fall into two categories: video compression that condenses visual content for efficient context utilization~\cite{li2024vidtome,lee2024video,shen2024tempme,shen2024longvu,choudhury2024don,zhang2024flash}, and chunk-based processing that segments long videos for step-wise inference~\cite{song2025moviechat+,tang2025adaptive,zhang2025silvr}. Integrated with powerful reasoning models, chunking solutions have evolved into agentic frameworks, dominating mainstream benchmarks with rational reasoning and tool scheduling~\cite{hurst2024gpt,guo2025deepseek,yang2025qwen3,fan2024videoagent,wang2024videoagent,wang2025videotree,feng2025video,tang2025video,qiu2026longvideo}. Targeting extreme ultra-long video scenarios, \benchname further motivates robust agentic designs and memory-based mechanisms to address long-range video understanding challenges~\cite{wang2025videollamb,diko2025rewind,jin2025videomem,zuo2025videolucy}.

\begin{figure}[!t]
\centering
\includegraphics[width=\linewidth]{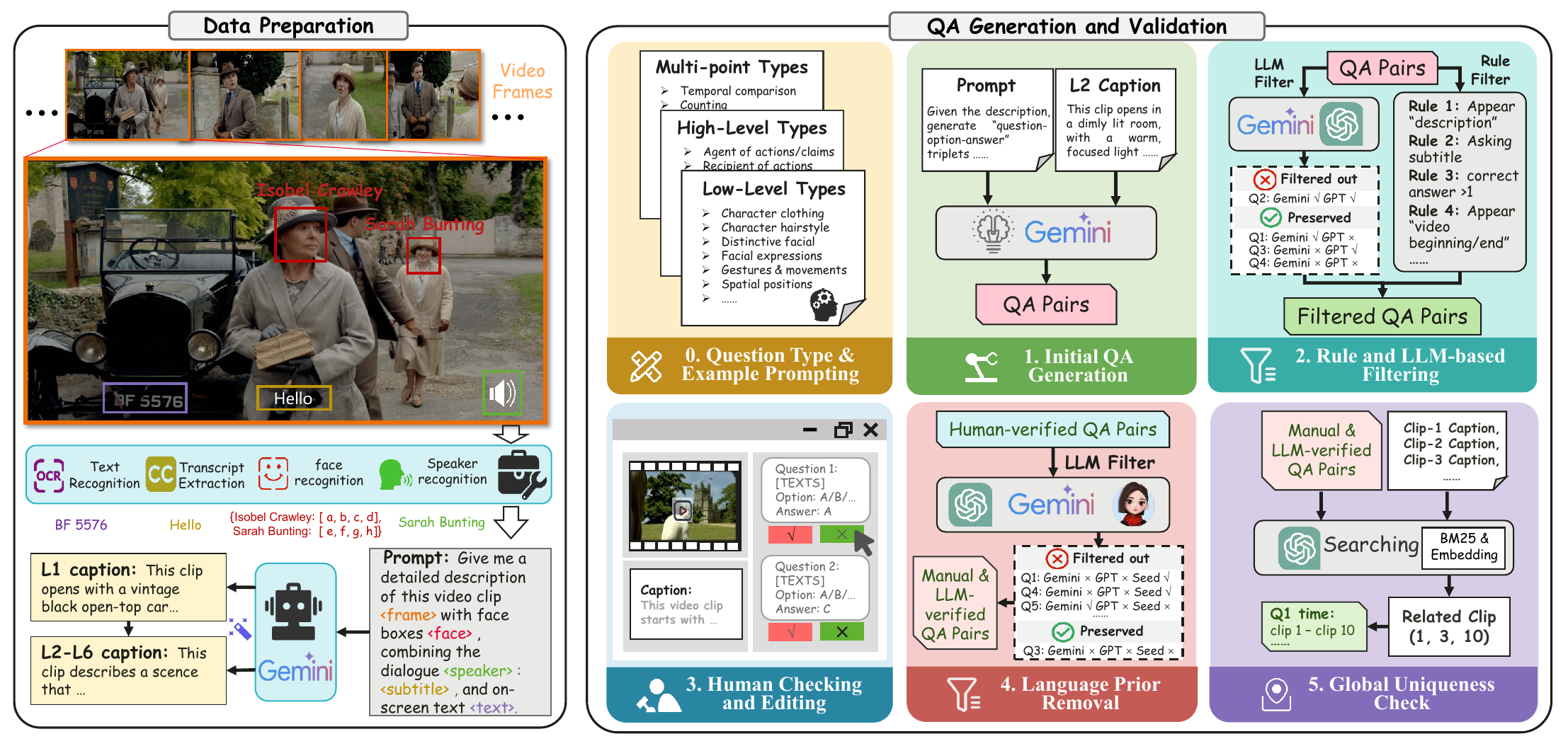}
\caption{An illustration of the two-stage procedure to establish the \benchname benchmark.} 
\label{fig:pipeline}
\end{figure}

\section{The \benchname Benchmark}
\label{benchmark}

The overall pipeline for building the \benchname benchmark is shown in Figure~\ref{fig:pipeline}. This process consists of two main stages: (1) data preparation, where we extract structured metadata from the TV dramas and use an MLLM to generate detailed video descriptions across different time levels; and (2) QA generation and validation, which combines LLM generation with strict human checking to ensure both the large scale and the high quality of the dataset.

\subsection{Data Preparation}
\label{benchmark:preparation}

\noindent\textbf{Data collection.}
Our dataset comprises 15 long‑form TV series spanning multiple seasons, with a total runtime of 606 hours. Specifically, \benchname includes \dramas totalling 466 hours, and the other three series are reserved for the training split.  We extract metadata through a four-stage pipeline. We process each drama $\mathcal{V}$ individually; for simplicity, denote $T$ as its total duration, and omit the drama index for the rest of this work. Details of this part can be found in Appendix~\ref{data_preparation}.


\begin{enumerate}
\item \textbf{Transcript and text extraction.}
We use PaddleOCR~\cite{cui2025paddleocr,cui2025paddleocrvl} followed by Qwen3-VL-32B~\cite{bai2025qwen3} to extract and refine hardcoded subtitles and in-screen texts. This produces a sequence of utterances $\mathcal{U}=\{u_1,\dots,u_N\}$, where each $u_n=(\mathbf{w}_n,[s_n,e_n])$ contains the text $\mathbf{w}_n$ and start/end timestamps $s_n,e_n\in[0,T]$, and a set of texts $\mathcal{T}=\{t_1,\dots,t_M\}$, where each $t_m=(\mathbf{w}_m,[s_m,e_m])$.
\item \textbf{Character library construction.}
We build a cast roster $\mathcal{C}$ from credits and web data. We initialize a facial registry with official portraits and enrich it using 1-FPS video sampling to capture varied appearances. Then, a frame-by-frame processing is performed to detect all facial instances $\mathcal{F}=\{f_1,\dots,f_R\}$. Here, each $f_r=(c_r,t_r,\mathbf{b}_r)$ links a detected face to an identity $c_r\in\mathcal{C}$ at timestamp $t_r$ and bounding box $\mathbf{b}_r$.
\item \textbf{Speaker label annotation.}
We combine audio features within $[s_n, e_n]$ and visual instances $\mathcal{F}$ to assign initial speaker labels using a label propagation algorithm to find `who spoke when'. Human labelers then perform a sentence-by-sentence check following a well-established guideline, updating each utterance $u_n$ with a final speaker identity $c_n\in\mathcal{C}$.
\item \textbf{Hierarchical video captioning.}
We use a multi-level temporal structure (L1--L6) to summarize the long video. First, the video is split into 10--15s base units (called L1 units) $h^{(1)}_k \in \mathcal{H}^{(1)}$ using the PySceneDetect algorithm~\cite{pyscenedetect} with CLIP similarity~\cite{radford2021learning}. For these units, Qwen3-VL-32B~\cite{bai2025qwen3} generates visual descriptions $\mathbf{d}^{(1)}_k$. Next, we cluster these into L2 units and use Qwen3-32B~\cite{yang2025qwen3} to summarize them. This repeats across six levels, $\mathcal{H}^{(l)}$ for $l \in \{1, \dots, 6\}$. Each segment is formalized as $h_k^{(l)}=(\mathbf{d}_k^{(l)},[s_k^{(l)},e_k^{(l)}])$, with start and end timestamps $s_k^{(l)}$ and $e_k^{(l)}$. This ends with a single L6 summary for the full drama $\mathcal{V}$.
\end{enumerate}

\subsection{QA Generation and Validation}
\label{benchmark:qa_generation}

We design a sophisticated pipeline to generate QA pairs from the multi-level captions, based on three main rules: (i) \textbf{diversity}, ensuring a wide coverage of question types; (ii) \textbf{reducing language prior}, making sure models cannot answer questions using only their internal knowledge (like drama scripts in their training data) without seeing the video; and (iii) \textbf{global uniqueness}, ensuring questions are not repeated across the entire drama to test ultra-long reasoning. To prevent the data from favoring any specific model, we use several different LLMs across the different stages. Details of all these steps are in Appendix~\ref{qa_generation}.

\begin{enumerate}
\item \textbf{Initial QA generation.} We use Gemini-2.5-Pro to generate initial QA pairs from L2 captions $h_k^{(2)}$. These segments, with an average length of 1.2 minutes, offer a good balance between visual details and the main storyline. Following previous methods (\textit{e.g.}, LLaVA-Video-178K~\cite{zhang2024llava}), we ask for different question types (with demonstrations) in the prompt, including \qtvp, \qtsu, \qtai, \qteu, \qttu, \qtrr, and \qtco. All entries are single-choice questions with 2--8 (average 6+) options.
\item \textbf{Rule and LLM-based filtering.} To reduce hallucination and internal knowledge prior, we first use an LLM (Qwen3-32B) to remove `self-answering' questions where the answer is already in the question text, as well as invalid questions that rely on segment-specific time clues (\textit{e.g.}, `at the beginning of the clip'). In addition, two LLMs (GPT-5 and Gemini-2.5-Pro) are asked to answer the questions based only on the text options; any QA answered correctly by both models is removed.
\item \textbf{Human checking and editing.} We recruit human labelers to check the remaining questions. Labelers are assigned L2 segments and asked to watch the video $h_k^{(2)}$ to make sure each QA meets three rules: (a) the question matches the video and is concise; (b) the choices are clear; (c) the answer is correct and unambiguous; and (d) the question is not repeated. We implement a two-stage labeling procedure, where the second pass rigorously double-checks and improves the quality of the first pass.
\item \textbf{Language prior removal.} After human editing, we give the QAs an even stricter text-only test using three models: GPT-5, Gemini-2.5-Pro, and Seed1.8 Pro. To further increase the difficulty, we only keep QA pairs that are answered incorrectly by all three LLMs. This high standard ensures a strong gap between text-only and vision-language results.
\item \textbf{Global uniqueness check.} We use a global search algorithm to make sure each QA is unique across the entire drama length $T$. We combine language embeddings from the bge-large-en/zh-v1.5 model~\cite{xiao2024c} and the BM25 algorithm~\cite{robertson2009probabilistic} to find similar L2 segments. Fo the top-ranked similar segments, an LLM extracts keywords from the QA; if another L2 segment contains these keywords, the QA is marked as not unique and removed. This ensures that the question can only be answered by referring to the specific target video segment, thereby improving the uniqueness and discriminative power of the QA pairs.
\end{enumerate}

The dataset contains 25,404 L2 segments, of which 20,278 belong to \benchname. After the first four steps, the average numbers of QA pairs per L2 segment are 103.1, 27.3, 18.8, and 5.4, respectively. The training set retains all QA pairs (about 15k) generated after Step 4. These pairs are then processed in Step 5 to annotate the uniqueness level of each question. The benchmark set retains only QA pairs that are globally unique within their entire drama after Step 5, resulting in 1.7 QAs per L2 segment. We then sample the final \fullQAs questions to ensure balanced distributions across question types and video timestamps.


\begin{figure}[!t]
\centering
\vspace{-0.4cm}
\includegraphics[width=0.98\linewidth]{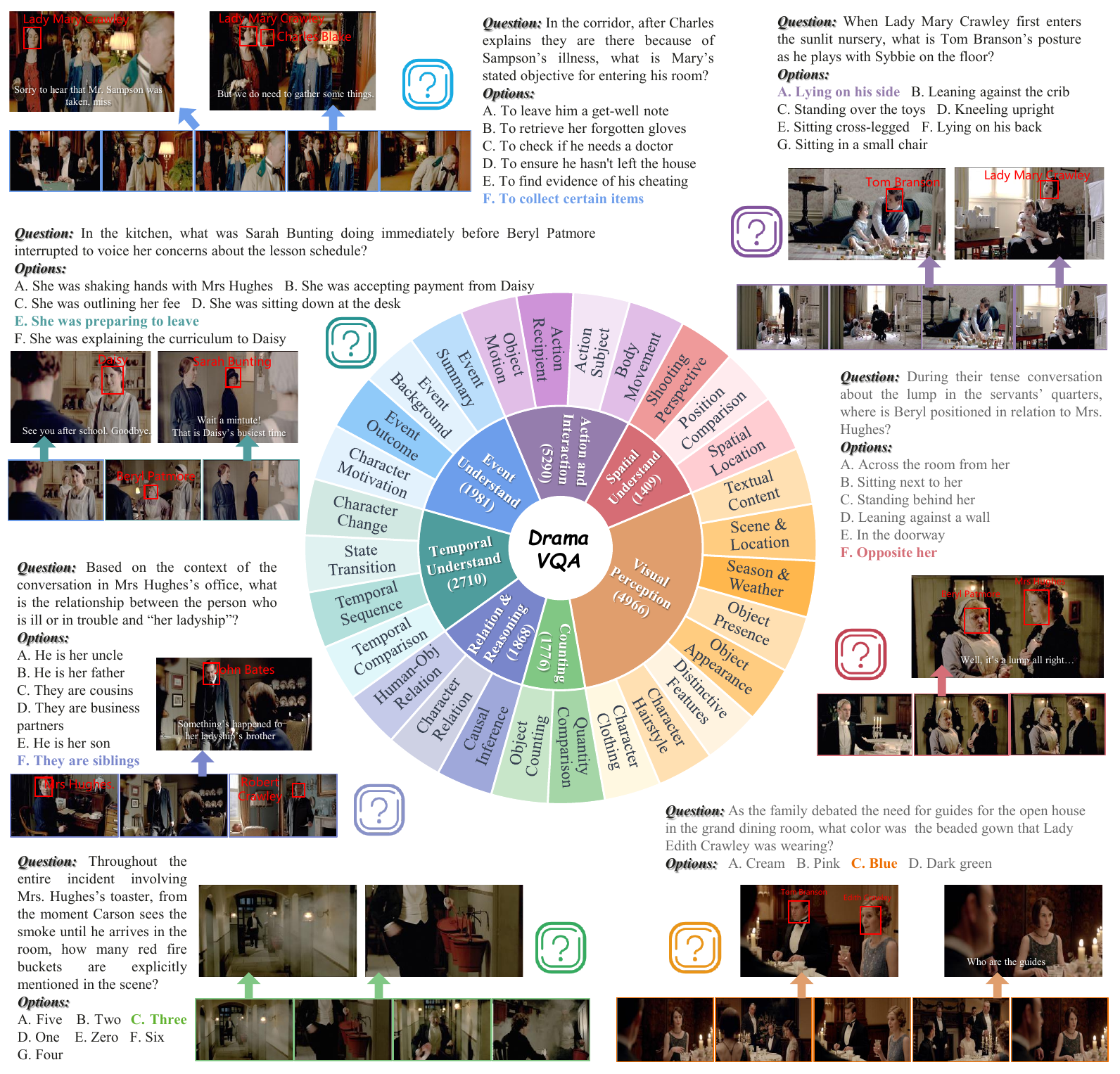}
\caption{Two-level QA categorization and representative examples (one for each Level-1 category; the color of each question logo corresponds to that of the category sector). We also display the keyframes that the model must sample for question-answering.} 
\label{fig:QA_examples}
\vspace{-0.4cm}
\end{figure}

\subsection{Benchmark Statistics}
\label{benchmark:statistics}

We have established \QAs QA pairs across \dramas TV dramas, representing a scale one order of magnitude larger than existing long-video benchmarks. Furthermore, the filtered QAs (400K+), while not unique throughout the entire drama, can also be used as training data and served as a high-quality resource for supervised fine-tuning and advancing long-form video understanding.
An illustration of seven basic categories and their breakdown into 28 sub-categories is shown in Figure~\ref{fig:QA_examples}. Detailed dataset statistics are provided in Appendix~\ref{data_preparation} (Table~\ref{tab:benchmark_statistics}).

Because each preserved question is verified to be unique throughout the entire drama duration $T$, \benchname supports \textbf{multi-granular evaluation}. In our experiments, we use L3 through L6 segments to define four distinct evaluation tracks: \tmedium (averaging 7.6 minutes), \tlong (averaging 43.1 minutes), \textralong (averaging 9.3 hours), and \tultralong (the full drama, averaging 38.9 hours). Notably, our \tmedium and \tlong tracks are similar in length to the `long-form' subsets of existing benchmarks, while the \textralong and \tultralong tracks present an unprecedented challenge for modern MLLMs.

Another interesting feature of our benchmark is the \textbf{high degree of inter-connectivity} between QA pairs within a single drama. For example, the answer to one question may serve as a condition for another. These connections enable us to construct complex tasks that require multi-hop reasoning or retrieving and integrating multiple pieces of information, providing a rigorous assessment of MLLMs’ reasoning and multi-point information retrieval capabilities. We present a preliminary study of these complex reasoning tasks in Section~\ref{experiments:riddles}.

\section{Experiments}
\label{experiments}

\subsection{Models and Evaluation Metrics}
\label{experiments:setting}

\noindent \textbf{Models.}
We evaluate two main groups of models to build a strong baseline. The first group includes state-of-the-art MLLMs, such as Qwen3-VL-8B/32B~\cite{bai2025qwen3}, Qwen3.5-27B~\cite{qwen35blog}, Qwen3.6-27B~\cite{qwen36plus}, GLM-4.6V~\cite{hong2025glm} and GLM-5.3-Flash-FP8~\cite{glm532026}. These models uniformly sample a set number of frames (256 or 512) from the input video, tokenize them as visual context, and then perform inference. We strictly follow the default settings for these models; Thinking settings are specified in Table~\ref{tab:main_results}. The second group features recently released, agentic MLLMs, namely Conan~\cite{ouyang2025conan} and LongVideo-R1~\cite{qiu2026longvideo}, which start with a basic video input and dynamically sample extra frames for question-answering.

Furthermore, we test a simple baseline named Naive-RAG that searches through the clip captions generated across the video. This method shares the same input structure as LongVideo-R1~\cite{qiu2026longvideo}: for each input video, we build a 3-level tree structure, where each clip is evenly divided into $D=6$ smaller clips at the next level. We extract a caption for every clip and use BM25 search to find the top-3 most relevant captions at each level. Finally, these 9 captions are combined and fed into an LLM to answer the question. We also develop a beam-search variant (RAG-BS) where, at the second and third levels, the search is strictly limited to the child nodes of the top-3 captions chosen in the previous level.


\noindent \textbf{Data.}
We test the models on the entire set of \QAs QA pairs created in the previous section. As explained earlier, each question is tested using different lengths of video input. These correspond to four distinct tracks: \tmedium, \tlong, \textralong, and \tultralong. To make it easier for the community to reproduce our results, we also build a `mini' subset containing about 5\% of the main test set. On this smaller set, we test tree proprietary models, Seed2.0-Pro~\cite{seed20modelcard}, GPT-5.5~\cite{openai_gpt55_2026} and Gemini-3.5-Flash~\cite{team2023gemini}, and compare them against the open-source models.



\noindent \textbf{Evaluation Protocol.}

For each sampled frame, original hardcoded subtitles are retained, forcing models to read text directly from images. To aid character recognition, we draw red bounding boxes for detected faces and add name labels. Models are required to output a single option (e.g., \texttt{[A]}). For each QA pair, we repeatedly prompt MLLMs for up to five attempts. If no valid answer is produced, we assign a random-guess score of $1/X$, where $X$ denotes the number of options.
We report both QA accuracy and computational overhead\footnote{All models run on an 8$\times$A800 GPU server with vLLM. Latency is measured in single-thread mode, while multi-thread testing brings 6--10$\times$ speedup.}. Notably, the overhead of agentic methods, including RAG, covers video captioning costs. Full experimental settings are provided in Appendix~\ref{experimental_results:setting}.

\begin{table}[!t]
\centering
\begingroup

\fontsize{6.4}{7.3}\selectfont
\setlength{\tabcolsep}{2.5pt}
\renewcommand{\arraystretch}{1.0}

\newcommand{\LVtick}{\ding{51}}
\newcommand{\LVcross}{\ding{55}}

\def\LVsplit#1/#2/#3\LVstop{%
  \makebox[2.2em][r]{#1}%
  \if\relax\detokenize{#2}\relax
  \else
    \,[#2]%
  \fi
  \,\textcolor{black!65}{%
    \fontsize{5.7}{6.5}\selectfont(#3)%
  }%
}
\newcommand{\LVCell}[1]{%
  \LVsplit#1\LVstop
}

\newcommand{\LVRow}[7]{%
  #1 & #2 & #3
  & \LVCell{#4}
  & \LVCell{#5}
  & \LVCell{#6}
  & \LVCell{#7} \\
}

\newcommand{\LVSection}[1]{%
  \rowcolor{lightgray}
  \multicolumn{7}{@{}l@{}}{%
    \strut\itshape #1%
  } \\
}

\begin{tabular*}{\linewidth}{@{\extracolsep{\fill}}lccllll@{}}
\hline
\textbf{Model}
& \textbf{Fr.}
& \textbf{Th.}
& \multicolumn{1}{c}{
    \shortstack{\textbf{Medium (L3)}\\Mini [Full] (time)}}
& \multicolumn{1}{c}{
    \shortstack{\textbf{Long (L4)}\\Mini [Full] (time)}}
& \multicolumn{1}{c}{
    \shortstack{\textbf{Extra-long (L5)}\\Mini [Full] (time)}}
& \multicolumn{1}{c@{}}{
    \shortstack{\textbf{Ultra-long (L6)}\\Mini [Full] (time)}} \\
\hline

Human & -- & --
& \LVCell{81.0//\textit{215}}
& \makebox[2.2em][r]{--}
& \makebox[2.2em][r]{--}
& \makebox[2.2em][r]{--} \\

\LVSection{No-vision Models}

\LVRow{Random Guess}{0.0}{\LVcross}
  {16.2/16.2/0.0}
  {16.2/16.2/0.0}
  {16.2/16.2/0.0}
  {16.2/16.2/0.0}

\LVRow{Language Only (Q3)}{256}{\LVcross}
  {21.5/22.2/0.1}
  {21.5/22.2/0.1}
  {21.5/22.2/0.1}
  {21.5/22.2/0.1}

\LVRow{Language Only (Q3.5)}{256}{\LVtick}
  {23.1/23.3/33.9}
  {23.1/23.3/33.9}
  {23.1/23.3/33.9}
  {23.1/23.3/33.9}

\hline
\LVSection{Proprietary Models}

\LVRow{Seed2.0 Pro~\cite{seed20modelcard}}
  {256}{\LVtick}
  {73.2//48.5}
  {49.2//63.4}
  {32.2//60.1}
  {30.7//49.2}

\LVRow{GPT-5.5~\cite{openai_gpt55_2026}}
  {256}{\LVcross}
  {67.2//64.6}
  {45.2//55.5}
  {29.9//81.0}
  {27.6//56.1}

\LVRow{Gemini-3.5-Flash~\cite{team2023gemini}}
  {256}{\LVcross}
  {77.5//38.5}
  {61.2//36.4}
  {43.9//36.8}
  {37.1//56.0}

\hline
\LVSection{Open-source Native MLLMs}

\LVRow{GLM-4.6V-106B~\cite{hong2025glm}}
  {256}{\LVcross}
  {51.0//15.4}
  {34.5//14.8}
  {27.4//14.3}
  {26.5//12.6}

\LVRow{GLM-4.6V-106B~\cite{hong2025glm}}
  {256}{\LVtick}
  {57.1//19.5}
  {36.6//18.5}
  {27.1//16.1}
  {25.4//15.1}

\LVRow{GLM-5.3-Flash-FP8~\cite{glm532026}}
  {256}{\LVtick}
  {70.5//31.7}
  {49.9//14.4}
  {34.9//14.3}
  {32.1//14.2}

\LVRow{Qwen3-VL-32B~\cite{bai2025qwen3}}
  {256}{\LVtick}
  {62.0//20.5}
  {37.9//18.3}
  {26.4//14.9}
  {24.9//16.8}

\LVRow{Qwen3.5-27B~\cite{qwen35blog}}
  {256}{\LVcross}
  {69.4//26.3}
  {47.1//16.6}
  {35.7//13.2}
  {34.4//12.2}

\LVRow{Qwen3-VL-8B~\cite{bai2025qwen3}}
  {256}{\LVcross}
  {40.9/42.3/17.9}
  {28.7/30.2/11.5}
  {24.8/25.1/13.0}
  {22.7/24.1/11.2}

\LVRow{Qwen3-VL-8B~\cite{bai2025qwen3}}
  {256}{\LVtick}
  {54.2/56.3/19.2}
  {36.5/37.7/17.6}
  {24.6/26.7/16.3}
  {23.4/24.9/12.6}

\LVRow{Qwen3-VL-32B~\cite{bai2025qwen3}}
  {256}{\LVcross}
  {51.0/51.3/22.6}
  {35.1/35.0/17.9}
  {27.8/27.4/18.2}
  {26.4/25.9/17.4}

\LVRow{Qwen3.6-27B~\cite{qwen36plus}}
  {256}{\LVcross}
  {69.9/67.1/50.5}
  {48.0/49.5/23.8}
  {35.2/36.9/22.5}
  {33.0/34.6/20.1}

\hline
\LVSection{Open-source Native MLLMs}

\LVRow{GLM-4.6V-106B~\cite{hong2025glm}}
  {512}{\LVcross}
  {55.0//28.0}
  {37.0//29.3}
  {27.4//31.1}
  {26.6//30.6}

\LVRow{GLM-4.6V-106B~\cite{hong2025glm}}
  {512}{\LVtick}
  {60.8//34.7}
  {40.1//34.6}
  {25.6//35.0}
  {24.6//37.5}

\LVRow{GLM-5.3-Flash-FP8~\cite{glm532026}}
  {512}{\LVtick}
  {75.4//28.2}
  {56.1//33.5}
  {36.1//31.3}
  {32.1//32.6}

\LVRow{Qwen3-VL-32B~\cite{bai2025qwen3}}
  {512}{\LVtick}
  {66.1//28.6}
  {43.7//33.3}
  {26.4//30.7}
  {24.1//21.3}

\LVRow{Qwen3.5-27B~\cite{qwen35blog}}
  {512}{\LVcross}
  {72.2//27.4}
  {51.0//27.6}
  {37.1//26.4}
  {33.9//23.8}

\LVRow{Qwen3-VL-8B~\cite{bai2025qwen3}}
  {512}{\LVcross}
  {41.7/42.5/23.3}
  {30.4/31.1/26.0}
  {24.1/25.4/26.8}
  {22.6/24.2/29.0}

\LVRow{Qwen3-VL-8B~\cite{bai2025qwen3}}
  {512}{\LVtick}
  {57.2/58.9/29.8}
  {40.0/41.0/31.5}
  {26.9/27.5/32.2}
  {25.0/25.2/26.1}

\LVRow{Qwen3-VL-32B~\cite{bai2025qwen3}}
  {512}{\LVcross}
  {54.8/53.3/33.3}
  {35.8/37.2/36.4}
  {28.2/27.5/37.6}
  {25.4/26.2/37.9}

\LVRow{Qwen3.6-27B~\cite{qwen36plus}}
  {512}{\LVcross}
  {72.5/70.8/41.4}
  {55.0/54.9/40.9}
  {37.9/38.6/36.8}
  {34.1/35.1/30.4}

\hline
\LVSection{Open-source Agentic MLLMs}

\LVRow{Conan~\cite{ouyang2025conan}}
  {256}{\LVtick}
  {48.8/50.1/63.3}
  {36.3/37.1/60.5}
  {29.7/30.9/59.7}
  {28.6/29.9/68.3}

\LVRow{Conan~\cite{ouyang2025conan}}
  {512}{\LVtick}
  {49.0/48.8/87.5}
  {37.7/37.9/79.5}
  {29.7/30.5/79.6}
  {29.6/29.5/84.1}

\LVRow{RAG-BS (Q3)}
  {var}{\LVcross}
  {47.4/47.1/180.4}
  {33.5/33.7/372.1}
  {25.8/25.7/408.2}
  {24.8/24.7/406.5}

\LVRow{Naive-RAG (Q3)}
  {var}{\LVcross}
  {51.7/50.4/292.9}
  {46.6/45.9/1664.5}
  {29.7/29.1/1955.5}
  {26.3/25.3/1955.0}

\LVRow{RAG-BS (Q3.8)}
  {var}{\LVcross}
  {56.4/56.0/180.5}
  {40.4/41.1/372.2}
  {32.1/32.7/408.3}
  {30.9/31.5/406.5}

\LVRow{Naive-RAG (Q3.8)}
  {var}{\LVcross}
  {60.1/59.8/293.0}
  {56.7/56.1/1664.6}
  {36.9/36.8/1955.5}
  {31.5/32.7/1955.0}

\LVRow{LongVideo-R1~\cite{qiu2026longvideo}}
  {var}{\LVtick}
    {62.8/63.7/70.1}
  {54.8/55.3/90.1}
  {34.7/34.0/98.3}
  {27.2/26.6/93.2}

\LVRow{LongVideo-R1 (post-tuned on our data)}
  {var}{\LVtick}
  {66.1/67.5/81.7}
  {60.0/61.4/113.7}
  {40.0/40.9/163.8}
  {30.8/30.5/172.0}

\hline
\end{tabular*}
\endgroup

\caption{%
QA accuracy (\%) and average Mini inference time on \benchname.
Each cell reports Mini accuracy, Full accuracy in square brackets
when available, and Mini time (seconds) in parentheses.
\textbf{Fr.}: sampled frames;
\textbf{Th.}: thinking enabled;
\textbf{var}: varying frame counts.
Times include retries (at most five attempts per question);
more frequent failures can make smaller models slower.
Q3, Q3.5, and Q3.8 denote Qwen3-VL-32B, Qwen3.5-27B,
and Qwen3.8-27B, respectively.
Proprietary models are evaluated only on Mini;
their timing may include network latency.
}
\label{tab:main_results}
\end{table}

\subsection{Main Results}
\label{experiments:main_results}

\noindent Table~\ref{tab:main_results} summarizes all test results. We first discuss the available results on the \textbf{full-set} that contains all \fullQAs QA pairs. The key observations are listed below.

\textbf{The foundational QA tasks remain challenging.} Even within the \tmedium track, the tested models leave substantial room for improvement. Specifically, Qwen3.6-27B~\cite{qwen36plus} achieves the highest reported full-set accuracy at 70.8\% with 512 frames, but its accuracy falls to 35.1\% on the \tultralong track. This difficulty comes from three main factors: (i) our data preparation pipeline adds dense information into the video captions, allowing us to create difficult questions that require deep reasoning and an understanding of complex character relationships; (ii) most questions have 6--8 highly competitive (yet clear) options; and (iii) we removed simpler questions (those solvable via language priors) through our strict text-only filtering step.

\textbf{Performance decreases as the input video duration increases.} The average accuracy on the \tlong track is noticeably lower than that on the \tmedium track, and performance on the \textralong and \tultralong tracks drops close to the random guess baseline. This trend matches our expectation: as the video length grows, the chance of a model successfully sampling the key frames containing the answer becomes much lower. This also explains why all models show improved accuracy when the context window increases from 256 to 512 sampled frames. However, even with 512 frames, the performance remains close to the random baseline in the \tultralong track.

\textbf{Stronger base MLLMs produce better results.} Under the same 512-frame, non-thinking setting, \tmedium accuracy increases from 42.5\% for Qwen3-VL-8B to 53.3\% for Qwen3-VL-32B~\cite{bai2025qwen3} and 70.8\% for Qwen3.6-27B~\cite{qwen36plus}. The corresponding \tultralong scores are 24.2\%, 26.2\%, and 35.1\%, respectively. These results support \benchname as a useful testbed for evaluating foundation models. However, the smaller absolute gains on the longest track indicate that stronger base models still face substantial limitations when processing extended video contexts.

\begin{figure}[!t]
\centering
\includegraphics[width=\linewidth]{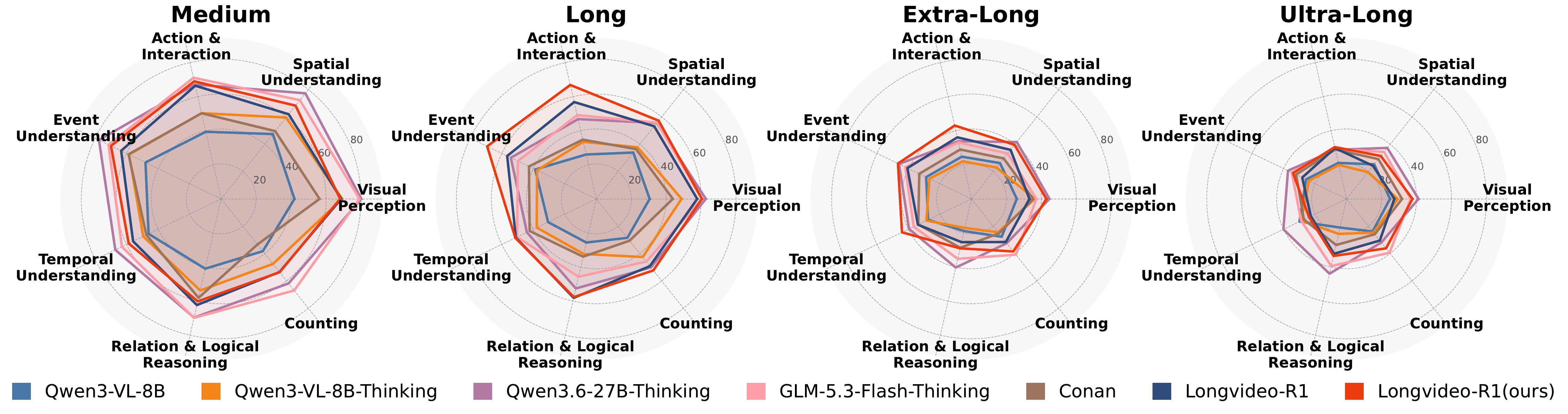}
\caption{Comparison of QA accuracy (\%) with respect to seven major QA categories.}
\label{fig:typewise_comparison}
\end{figure}

\textbf{Thinking mechanisms help, but their benefits depend on video length.} For Qwen3-VL-8B with 512 frames, thinking raises \tmedium accuracy from 42.5\% to 58.9\% and \tlong accuracy from 31.1\% to 41.0\%. However, the gain on the \tultralong track is smaller, from 24.2\% to 25.2\%. Similar trends hold with 256 frames, where thinking yields larger improvements on the shorter tracks. These results suggest that reasoning helps models interpret evidence, while offering limited benefits when sparse sampling fails to capture the information required to answer a question.

\textbf{Agent-based and retrieval-augmented algorithms show better accuracy but may require higher computational costs.} The basic RAG approach, which searches across all short-clip captions, helps reduce the information loss caused by uniform frame sampling. However, this comes at the cost of increased time complexity, which grows with the input video length. Smarter agentic models (\textit{e.g.}, LongVideo-R1~\cite{qiu2026longvideo}) offer a better approach by dynamically sampling a small number of video clips. Still, they cannot always guarantee finding the exact time segments needed for the correct answer. This shows a clear trade-off between accuracy and computational cost, which we discuss further in Section~\ref{experiments:tradeoff}.

\noindent \textbf{Results on the \textbf{mini-set}}. Its performance trends closely align with the full dataset, validating the mini-set as a reliable, cost-effective proxy for fast experiments such as parameter tuning and prompt design. Even proprietary models struggle with ultra-long contexts: while Gemini-3.1-Pro outperforms open-source native MLLMs, its accuracy still declines markedly on \tlong and \tultralong tracks, mainly due to severe information loss from sparse frame sampling across extended videos.

\noindent \textbf{RL on our training set}.
To assess the effectiveness of our training data in improving information retrieval and localization across video scales, we adopt the agent-based LongVideo-R1~\cite{qiu2026longvideo} as our baseline. Starting from its original checkpoint, we perform reinforcement learning on our training set. The resulting model outperforms the original LongVideo-R1 by at least 3.6\% across all four evaluation tracks, demonstrating the effectiveness of our training data. RL training increases caption retrieval and reasoning activity, with average medium-caption calls rising from 2.38 to 4.65, low-caption calls from 3.12 to 4.66, and reasoning steps from 7.41 to 10.94 per question.




\noindent \textbf{Summary of Table~\ref{tab:main_results}.}
With extensive experiments, we show that ultra-long video understanding remains a major challenge, especially under limited computational budgets. By showing the current limits of existing models, \benchname offers a valuable testbed to help guide the future development of state-of-the-art MLLMs and agentic video understanding algorithms.

\noindent \textbf{On QA categories.}
Figure~\ref{fig:typewise_comparison} shows the accuracy distribution across different question categories. Notably, certain question categories (\textit{e.g.}, \qttu and \qtco) are more difficult than others (\textit{e.g.}, \qtvp and \qteu). This difference happens mainly because the harder categories require complex logical reasoning or collecting pieces of information scattered across the video. See Appendix~\ref{experimental_results:detailed_results} for further analysis.

\noindent \textbf{Failure analysis.}
Our analysis reveals three main causes for these errors. First, many mistakes happen because sparse frame sampling often misses important keyframes. Second, current models struggle to understand the overall story logic and cause-and-effect relationships. Finally, the lack of audio inputs makes it hard for models to answer dialogue-based questions, such as identifying `who claimed what'. This clearly shows the need to develop advanced tri-modal (audio-visual-text) MLLMs in the future. More examples are provided in Appendix~\ref{experimental_results:qualitative_studies}.

\subsection{Accuracy-Efficiency Tradeoff}
\label{experiments:tradeoff}

We compare native MLLMs and agentic algorithms from the perspective of the accuracy-efficiency trade-off. In brief, while agentic methods introduce higher preprocessing cost in case-wise evaluation, they become substantially more competitive in batch-wise settings by sharing captioning cost across many questions from the same long video context. We also observe a practical trade-off between stronger reasoning performance and additional inference overhead. Detailed quantitative results, the full comparison table, and time-cost estimation are provided in Appendix~\ref{tradeoff}.

\subsection{Multi-Point and Multi-Hop Reasoning: A Preliminary Study}
\label{experiments:riddles}

We extend \benchname with multi-point and multi-hop questions
composed from basic QA pairs to evaluate cross-segment evidence
integration and compositional reasoning.

\paragraph{Question construction.}
Multi-point questions cover
\textit{Two-/Three-Event Temporal Comparison} (TTC-2/3) and
\textit{Cross-Event Character Identity} (CECI).
We sample events from distinct L3 segments within a drama
and use templates to query their temporal order or whether
characters identified by two base questions are the same.
An LLM refines the wording, and we discard questions answered
correctly by at least two of three text-only LLMs.
For multi-hop questions, we pair questions whose first answer
(typically a character name) appears in the second question.
An LLM replaces this answer in the second question with a
description derived from the first; annotators verify correctness
and the necessity of the intermediate answer.
See Appendix~\ref{riddles} for detailed construction procedures,
illustrative examples, and supplementary experimental analyses.

\paragraph{Evaluation.}
We uniformly sample 256 frames from the concatenated L3 segments
associated with each composed question.
Table~\ref{tab:multi_point_multi_hop_results} shows that
Qwen3-VL and GLM-4.6V perform near or below chance on TTC-2
and CECI, and only slightly above chance on TTC-3.
Despite stronger temporal comparison by Qwen3.5/3.6,
CECI accuracy peaks at 59.0\%.
All models perform worse on multi-hop questions than on
their paired second-hop questions, highlighting reasoning
challenges beyond atomic QA.
Together, these results reveal limitations in cross-segment
evidence integration and compositional reasoning that atomic
QA evaluation alone may not fully capture.

\begin{table}[!t]
    \centering
    \small
    \renewcommand{\arraystretch}{1.2}
    \setlength{\tabcolsep}{7pt}

    \resizebox{0.68\linewidth}{!}{%
    \begin{tabular}{@{}lcccccc@{}}
        \toprule
        \multirow{2}{*}{\textbf{Model}}
        & \multicolumn{4}{c}{\textbf{Multi-Point (1,510)}}
        & \multicolumn{2}{c}{\textbf{Multi-Hop (1,200)}} \\
        \cmidrule(lr){2-5} \cmidrule(l){6-7}
        & \textbf{TTC-2}
        & \textbf{TTC-3}
        & \textbf{CECI}
        & \textbf{Average}
        & \textbf{Single-QA}
        & \textbf{Multi-Hop-QA} \\
        \midrule
        Random
        & 50.0 & 16.7 & 50.0 & 39.6 & 13.7 & 16.0 \\
        Qwen3-VL-8B~\cite{bai2025qwen3}
        & 47.4 & 18.6 & 49.2 & 38.2 & 39.5 & 38.4 \\
        Qwen3-VL-32B~\cite{bai2025qwen3}
        & 46.4 & 19.9 & 43.4 & 36.1 & 44.8 & 42.2 \\
        GLM-4.6V-106B~\cite{hong2025glm}
        & 47.4 & 18.8 & 48.2 & 37.8 & 48.4 & 46.0 \\   
        Qwen3.5-27B~\cite{qwen35blog}
        & 66.3 & 57.9 & 50.6 & 57.3 & 61.0 & 56.8 \\
        Qwen3.6-27B~\cite{qwen36plus}
        & \textbf{81.1} & \textbf{70.9} & \textbf{59.0} & \textbf{69.0} & \textbf{67.4} & \textbf{63.1} \\        
        \bottomrule
    \end{tabular}%
    }
\vspace{-5pt}
    \caption{Evaluation on multi-point and multi-hop questions.
\textit{Single-QA} denotes the base question taken from the second hop of the paired \textit{Multi-Hop-QA} sample, with identical video input. Parenthesized values are the total number of questions per task.}
\vspace{-10pt}
    \label{tab:multi_point_multi_hop_results}
\end{table}

\section{Conclusions}
\label{conclusions}
\vspace{-10pt}

In this paper, we introduced \benchname, a large-scale benchmark designed to advance ultra-long video understanding. Built from long TV drama series, \benchname contains about \QAs high-quality QA pairs and evaluates models from \tmedium clips to \tultralong videos lasting tens of hours. Our results show that current MLLMs still struggle on \textralong and \tultralong tracks, revealing clear limitations in long-range video reasoning.
We also explored the accuracy-efficiency trade-off, finding that agent-based algorithms offer a better balance when answering many questions over long videos. Moving forward, we believe future research must shift from simply expanding context windows toward developing smart memory systems and video-driven world models that can efficiently manage and reason over long-term temporal contexts.

\bibliographystyle{plainnat}
\bibliography{main}

\newpage
\appendix

\section{Data Preparation Details}
\label{data_preparation}

\subsection{Video Pre-processing}
\label{data_preparation:video_preprocessing}

Each drama in our dataset contains multiple episodes. To ensure the story flows naturally, we first remove the opening and closing parts (titles and credits) of each episode. The remaining video parts are combined into a single, continuous video stream for each drama\footnote{We do not actually perform video concatenation, but we compute a unique index for each frame (across all curated episodes), so that when we want to sample a number of frames, we can just compute the list of frame index and fetch image/audio data from the corresponding video file.}. This step provides a longer time context for conversations that happen near episode boundaries. It helps the \modelname keep character consistency and identify speakers based on the overall story.

\subsection{Transcription Extraction}
\label{data_preparation:transcription}

We extract dialogue text from on-screen subtitles using an OCR and refinement pipeline.
\begin{itemize}
\item \textbf{Initial subtitle extraction.} For each drama, we define a box around the subtitle area to reduce background noise. We then use PaddleOCR-v4~\cite{cui2025paddleocr,cui2025paddleocrvl} for frame-by-frame recognition, producing a raw text sequence $\{\mathbf{t}_n\}$.
\item \textbf{Grouping texts into utterances.} To combine continuous frames into separate dialogue lines, we calculate the normalized Levenshtein similarity (also known as $1$ minus edit distance) between adjacent frames:
$$\mathrm{sim}(\mathbf{t}_n,\mathbf{t}_{n+1})=1-\frac{\mathrm{Levenshtein}(\mathbf{t}_n,\mathbf{t}_{n+1})}{\max(|\mathbf{t}_n|,|\mathbf{t}_{n+1}|)}.\label{eq:subtitle_sim}$$
Frames with a similarity score higher than $0.8$ are grouped into a single utterance.
\item \textbf{MLLM refinement.} For each segment, the middle frame is processed by Qwen3-VL-32B~\cite{bai2025qwen3}. This step fixes OCR errors, separates overlapping subtitles, and makes the text format consistent.
\item \textbf{Quality control.} We manually check about $1\%$ of the data, focusing on cases where the text length does not match the video duration or where the logical flow seems broken.
\end{itemize}

\subsection{Character Library Construction}
\label{data_preparation:character}

To build high-quality facial reference data, we use a two-stage pipeline that combines external and \textit{in-situ} visual data:
\begin{itemize}
\item \textbf{Initial face library.} Actor details and reference photos are collected from public databases (\textit{e.g.}, IMDB and TMDB for English dramas, Douban and BaiduBaike for Chinese dramas). We manually check the data to clearly map actors to their roles in the drama.
\item \textbf{\textit{In-situ} library refinement.} To handle changes in lighting, aging, and makeup within the drama, we scan the entire video at 1 FPS. Using the initial library as a reference, we extract all matching face images. To manage the size and quality of this data, we apply a step-by-step filtering process:
\begin{itemize}
\item \textbf{Multi-face exclusion.} Samples containing multiple faces are removed to avoid confusion.
\item \textbf{Deduplication.} We remove duplicate samples using a cosine similarity threshold of $0.8$.
\item \textbf{Quality filtering.} Low-resolution or blocked faces are removed based on a quality threshold of $0.45$.
\item \textbf{Quality control.} From the remaining high-quality pool, we randomly sample $15$ typical images per character to form the final reference library.
\end{itemize}
\item \textbf{Frame-wise scan.} Finally, we use the InsightFace library~\cite{deng2019arcface} to perform frame-by-frame detection and recognition across the full length of each drama, providing the visual data needed to identify speakers.
\end{itemize}

\subsection{Speaker Label Initialization and Annotation}
\label{data_preparation:speaker}

\noindent \textbf{Acoustic feature extraction.}
We use the ffmpeg library to extract raw audio from the videos at a sampling rate of $16\mathrm{kHz}$. For each dialogue segment $n$ in the set $\mathcal{S}=\{(\mathbf{t}_n,a_n,b_n)\}_{n=1}^N$, we cut out the matching audio clip using the pydub library. Voice features are created using the ERes2Net model~\cite{chen2023enhanced} within the \textbf{3D-Speaker} toolkit~\cite{chen20253d}. To make the similarity scores stable, all feature vectors are $\ell_2$-normalized. This maps them to a space where cosine similarity equals the dot product.

\noindent \textbf{Seed cluster generation.}
Given the possible utterance sets $\{\mathcal{S}_p\}$ linked to character $p$ based on nearby video frames, we start a strict clustering process. This is controlled by a similarity threshold $\theta_\mathrm{seed}$ (starting at $0.85$, a high value to ensure that two very similar utterances belong to the same person) and a minimum cluster size $\eta_\mathrm{seed}$. We then use a two-step search method:
\begin{itemize}
\item \textbf{Phase 1: Global high-confidence search.} For all active characters, we build an audio graph. An edge connects two nodes (utterances) if their similarity is above $\theta_\mathrm{seed}$. We find the maximum connected component (MCC) for each character using a floodfill algorithm. We rank these groups by size. If the largest group is big enough (meets $\eta_\mathrm{seed}$) and is different enough from existing seeds (similarity difference is above $\Delta\theta_\mathrm{seed}$), we accept it. If no group is accepted, we gradually lower $\theta_\mathrm{seed}$ by $0.01$ at a time until it reaches $0.70$.
\item \textbf{Phase 2: Targeted character search.} If any character $p$ still lacks a seed after Phase 1, we do a local search within $\mathcal{S}_p$. We use the same rules to check against existing groups to avoid mixing up identities. We slowly lower the threshold $\theta_\mathrm{seed}$ until every active character has one seed voice set.
\end{itemize}
This process includes an optional human check to make sure the seed sets are correct. For a typical drama with about $50$ characters, this guided process takes about 2 to 3 hours.

\noindent \textbf{Iterative affinity propagation.}
After finding the seeds, we spread the labels to the whole dataset using a step-by-step process. Starting from a high threshold $\theta_\mathrm{prop}$, we repeat a two-pass search:
\begin{itemize}
\item \textbf{Identity expansion.} Each unlabeled utterance is compared to all existing character groups. If the highest similarity is above $\theta_\mathrm{prop}$, the utterance is assigned to that character and added to the reference set. We then update the center of the group.
\item \textbf{Latent identity discovery.} Within a sliding window of $20$ dialogue lines, we search for groups using a stricter threshold ($\theta_\mathrm{prop}+\Delta\theta_\mathrm{prop}$). Groups larger than $\eta_\mathrm{prop}$ become new clusters for minor characters\footnote{Although these minor clusters act as unique identities and can spread labels to other utterances, they do not have visual references. To avoid mixing identities, we limit the spread of these labels to a small time window around the original group. This is different from main characters, who have visual anchors across the whole video.}. This lets the model track consistent voices missed by the first visual checks.
\end{itemize}
We lower $\theta_\mathrm{prop}$ by $0.02$ in each step. The process stops at $\theta_\mathrm{prop}=0.45$. Any utterances still unlabeled at this stage get the $\mathtt{[UNKNOWN]}$ tag, waiting for a final check by the LRM.

\noindent \textbf{Human annotation.}
Human labelers use a graphical user interface (GUI) that gives them three main types of information: (1) video streams with boxes drawn around detected faces, (2) initial speaker guesses from our algorithm, and (3) audio statistics (\textit{e.g.}, voice similarity scores). Labelers watch the videos in order and correct the initial labels where needed.

\begin{figure}[!h]
\centering
\fbox{
\begin{minipage}{0.9\linewidth}
\textbf{Guideline for human labelers: annotation rules and structure.} For each dialogue segment, the labeler must first group the utterance based on the number of active speakers.

\begin{enumerate}
\item \textbf{Single-speaker utterances.}
\label{rule_single}
\begin{enumerate}
\item \textbf{Main characters.} If the speaker is in the character library, the labeler must use their standard name.
\item \textbf{Minor or new characters.} If the speaker can be identified but is not in the library, the labeler must follow this naming order:
\begin{enumerate}
\item \textbf{Names.} Use full or partial names if spoken in the video (\textit{e.g.}, \textit{Jack} or \textit{Mr. Smith}).
\item \textbf{Job or social role.} Use their guessed job or social status (\textit{e.g.}, \textit{Doctor}, \textit{Director}).
\item \textbf{Off-screen sources.} For voices coming from off-screen, write a clear source description (\textit{e.g.}, \textit{Reporter}, \textit{Digital Voice}). Use specific markers like \textit{Narration} or \textit{Interlude} when needed.
\item \textbf{Relationships.} Use family or social links to main characters (\textit{e.g.}, \textit{Alice's Daughter}).
\item \textbf{Visual looks.} If no other details are known, use obvious physical features (\textit{e.g.}, \textit{Man in Black}, \textit{Short-haired Woman}).
\item \textbf{Additional: numerical indexing.} Separate similar unknown characters using numbers (\textit{e.g.}, \textit{Worker \#1}, \textit{Worker \#2}, \textit{etc.}).
\end{enumerate}
\item \textbf{Unknown.} Use the $\mathtt{[UNKNOWN]}$ tag if the person is still unclear after checking both audio and video.
\end{enumerate}
\item \textbf{Simultaneous multi-speaker utterances.}
\begin{enumerate}
\item \textbf{Small groups.} If $P'\leqslant5$ speakers can be identified, write down all their names.
\item \textbf{Large crowds.} If there are more than $5$ speakers or the speech is a chorus, use the $\mathtt{[MULTIPLE]}$ tag. This can be mixed with specific names if main characters can be clearly heard in the crowd (\textit{e.g.}, \textit{Alice} and \textit{Bob} and $\mathtt{[Multiple]}$ \textit{Students}).
\end{enumerate}
\item \textbf{Sequential composite utterances.} For quick interruptions or lines with multiple speakers speaking one after another, the labeler must split the text and label each part separately.
\end{enumerate}
\end{minipage}
}
\caption{A brief version of the annotation guideline for labelers.}
\label{fig:annotation_guideline}
\end{figure}

To ensure high-quality labels, especially for minor roles without early voice samples, we use a strict quality check process. After the first round of labeling, a computer script finds uncertain labels for a second human check. A label is marked for review if it meets any of these rules: (a) a main character is assigned to a voice that sounds very different from their normal voice, (b) a main character speaks while off-screen, (c) a new minor character appears, or (d) the tag is $\mathtt{[UNKNOWN]}$.

On average, the first labeling step takes 1.5 to 2.5 hours per video hour, while the targeted quality check takes an extra 10 to 15 minutes per video hour. This multi-step approach reduces mistakes in the final data, especially for the hard cases discussed in Section~\ref{experiments}.

\subsection{Hierarchical Video Captioning}
\label{data_preparation:captioning}

\begin{figure}[!h]
\centering
\fbox{
\begin{minipage}{0.9\linewidth}
\textbf{The video clip description prompts} \\

    \promptsection{Task}
    You are a video understanding expert. Please create a description for the current video clip (containing multiple frames arranged in chronological order).

    \promptsection{Clip Description Guidelines}
    \begin{enumerate}
        \item \textbf{Character Description Guidelines}:
        \begin{enumerate}
            \item When referring to a character, use the name displayed above the red box around the character's face in the video frames. However, do not directly mention terms like `red box' in the description.
            \item If a character has no bounding box or name above the face, distinguish and describe them based on their appearance features, such as clothing style or other prominent characteristics.
            \item When a character appears for the first time, describe their features, such as clothing and hairstyle.
        \end{enumerate}
        \item \textbf{On-screen Text Description Guidelines}:
        \begin{enumerate}
            \item If text appears in the video frames, describe the text in its original language. If the recognized result is subtitles, it needs to be aligned with the additionally provided dialogue information.
            \item If there are objects with large sections of text in the frame, such as letters or notices, add the symbol $\langle\mathtt{texts}\rangle$ in the output description. For example: `He is holding a letter $\langle\mathtt{texts}\rangle$, which says...'.
        \end{enumerate}
        \item \textbf{Other Detail Description Guidelines}:
        \begin{enumerate}
            \item \textbf{Note}: Please provide as many details as possible, including the characters' movements, behaviors and expressions, interactions between characters, as well as details like the color and texture of the objects being manipulated.
            \item \textbf{Note}: The description can include the location where the story takes place, environmental details, weather characteristics, and important objects that reflect the background of the era.
        \end{enumerate}
    \end{enumerate}

    \promptsection{Reference Example}
    Assume the input information includes multiple video frames. \\
    Output Description: \{`Clip Level Description': `The clip depicts an outdoor scene on a cold winter day, on a dilapidated street with thick snow hanging from the eaves. The young boy Guangming is wearing XXX clothes and an XXX hat, with his cheeks flushed from the cold and snot dripping down. Zhou Bingkun is dressed in blue XXX clothes and an XXX hat, with a serious expression. The young boy Guangming pulls him, and Zhou Bingkun turns around and stops.'\}

    \promptsection{Output Format:}
    Your response should be in the following format: \{`Clip Level Description': `The clip XXX'\}
\end{minipage}}
\caption{The prompt to create detailed chronological descriptions for video clips based on multi-frame visual information with specific character and detail description guidelines.}
\label{fig:video_captioning_prompts}
\end{figure}

\begin{figure}[!h]
\centering
\fbox{
\begin{minipage}{0.9\linewidth}
\textbf{The video summarization prompts (datailed version)} \\
You are an expert in understanding descriptions of video clips, where the video consists of \{$\text{number\_of\_sub}$\} clips. Your task is to summarize these clip descriptions in chronological order to create a coherent video section description.

    \promptsection{Detailed Description Guidelines for Video Section Description}
    \begin{enumerate}
        \item Since clip descriptions are provided in chronological order, ensure the description is coherent and follows the same sequence. Do not refer to the first/last frame of any individual clip as that of the entire video.
        \item The clips are continuous; pay attention to maintaining logical coherence when summarizing.
        \item \textbf{Note:} Present text and dialogue from the clips in a paraphrased summary form. Avoid frequent use of expressions like `XXX said: `...''; instead, prefer `XXX did something and expressed XXX meaning'.
        \item \textbf{Note:} Due to clip segmentation, errors may exist. Correct such errors when merging clip descriptions, ensuring smooth narration at the junctions—for example, merge statements from the same person or descriptions of the same person/object. If subtitles for the entire section are provided, use their context to make judgments; only correct errors when confident, and do not make unfounded assumptions.
        \item \textbf{Note:} Merge repeated descriptions of a character's expressions, appearance, or state (e.g., `appearing helpless and powerless', `wearing a blue coat with a red badge on the chest', `continuing to wash dishes on the other side of the room without participating in the conversation') to avoid redundancy. Merge duplicated dialogue only if it is described twice for the same clip (not if the character actually spoke twice).
        \item \textbf{Note:} The tone of the video description should mimic direct narration of a video, not a summary of information from multiple clips. Thus, avoid expressions from the reference clip descriptions such as `this clip begins...', `as the clip progresses...', `this clip ends', `the final/initial frame', `the second clip starts with...', `the last few frames of this part'.
        \item \textbf{Note:} Incorporate all details from the given clip descriptions, but avoid repeating descriptions of the same shot. Try to understand the video's theme and provide a coherent narrative that connects all clips.
        \item \textbf{Note:} If subtitles or character relationships for these clips are provided, use them to aid understanding and correct errors.
        \item \textbf{Note:} Since the duration of each clip varies, the length of the concise description must meet the specific word count requirement for each case. Retain as much information as possible while reducing detailed descriptions.
        
    \end{enumerate}
    \promptsection{Output Format:}
    Your response must follow the following structure: 
    \{`Section Detailed Description': `The section ......'\}
\end{minipage}}
\caption{Video summarization prompts for consolidating individual clip descriptions into a unified, detailed narrative of the entire section.}
\label{fig:video_summarization_prompts}
\end{figure}

\begin{figure}[!h]
\centering
\fbox{
\begin{minipage}{0.9\linewidth}
\textbf{The video summarization prompts (From the detailed version to the brief version).} \\
You are an expert in understanding video clip descriptions. Your task is to summarize a concise description of the video based on its detailed description, and generate a corresponding title.

    \promptsection{Guidelines for Concise Description}
    \begin{enumerate}
        \item The concise description is a summary of the detailed description.
        \item It must include key elements such as people/objects involved, actions performed, locations, and core events. Retain information useful for understanding the plot, but omit excessive detailed descriptions.
        \item It should contain distinguishing features of the scene, such as the story's setting, unique plot points, or main character relationships.
        \item If subtitles or character relationships of the video are provided, use them to check and correct any errors.
        \item \textbf{Note:} Retain as much key information as possible while minimizing redundant details.
    \end{enumerate}

    \promptsection{Guidelines for Title}
    \begin{enumerate}
        \item The title should be in the form of a phrase, concisely capturing the core event of the video.
    \end{enumerate}

    \promptsection{Output Format:}
    Your response must follow the following structure: 
    \{`Section Brief Description': `The section XXXX', `Title': `XXX' \}
\end{minipage}}
\caption{The prompt to condense the description of the entire video segment into a concise version.}
\label{fig:video_summarization_prompts_brief}
\end{figure}

\noindent \textbf{Clip segmentation via adaptive merging.} We use the PySceneDetect library to cut the video into basic shots based on visual content. By using an adaptive threshold algorithm that checks changes in lighting and color, we divide the video into separate shots with an average length of $2$ to $3$ seconds.

To group these shots into meaningful clips, we use a multi-view feature method. For each shot, we extract the first, middle, and last frames and calculate their features using the CLIP ViT-L model~\cite{radford2021learning}. Joining these three vectors creates a strong representation of the shot's visual content. We then calculate the cosine similarity between neighboring shots. To handle different visual styles across dramas, we use a dynamic threshold: the merging threshold is set to the top $1/5$ value of all similarity scores within a specific drama. Neighboring shots with scores above this threshold are merged until they reach a target length of $8$ to $15$ seconds. This step makes sure the clips are long enough to show the story but short enough for the multimodal model to process easily.

\noindent \textbf{Base clip captioning.} Let the generated clip sequence be $\mathsf{V}=(\mathsf{C}_1,\ldots,\mathsf{C}_L)$. For each clip $\mathsf{C}_l$, we sample frames at 1 FPS, focusing on frames with many faces. We represent each clip as a sequence of these sampled frames and use Qwen3-VL-32B~\cite{bai2025qwen3} to generate captions. To improve the accuracy of these captions, we use two methods to link text and visuals: (a) Text adding: We add the dialogue text and current speaker guesses directly into the model's input context (if the text comes from outside files, \textit{e.g.}, an SRT file, we draw the text onto the matching video frames), and (b) Visual boxes: We draw red boxes around detected faces, labeled with the matching character ID. This produces detailed descriptions averaging $300$ words per clip, ensuring that the visual story is closely tied to character identities. Figure~\ref{fig:video_captioning_prompts} shows the prompts for video clip captioning.

\noindent \textbf{Hierarchical Segment Summarization.} To handle the high cost of processing millions of words per drama, we group clips into continuous sections $\mathsf{S}_m$ based on meaning. We extract text features for each clip caption using the BAAI bge-large-zh-v1.5 model~\cite{xiao2024c}, and calculate the cosine similarity between neighboring clips $(\mathsf{C}_l,\mathsf{C}_{l+1})$. Clips are merged into a section if their caption similarity is over $0.8$, with a limit on the section size (around $8\pm4$ clips).

For each created section $\mathsf{S}_{l'}$, we use Qwen3-32B~\cite{yang2025qwen3} to combine the clip captions into a single, high-level segment summary $\mathbf{T}_{l'}$. This summary removes repeated visual details and focuses on main plot events and character interactions. Figure~\ref{fig:video_summarization_prompts} shows the prompts for video segment summarization.

\subsection{Dataset Statistics}
\label{data_preparation:statistics}

Table~\ref{tab:benchmark_statistics} provides detailed statistics of \benchname across four evaluation granularities. As the temporal level increases from \tmedium to \tultralong, each video covers a substantially longer context and contains more QA pairs. The table also reports metadata scale, including facial bounding boxes, speaker labels, subtitle length, caption length, and average QA length, highlighting the dense multimodal annotations provided by our benchmark.

\begin{table}[!t]
\centering
\resizebox{\linewidth}{!}{
\begin{tabular}{|l|c|c|c|c|c|c|c|c|}
\hline
\textbf{Benchmark} & \textbf{\#Vids} & \textbf{Dura.} & \textbf{\#QAs} & \textbf{\#Faces} & \textbf{\#Spkrs} & \textbf{Sub Len.} & \textbf{Cap Len.} & \textbf{QA Len.} \\
\hline
Video-MME-S & 300 & 0.02h & 3.0 & N/A & N/A & 198.6 & N/A & 28.7 \\
Video-MME-M & 300 & 0.14h & 3.0 & N/A & N/A & 1425.6 & N/A & 32.8 \\
Video-MME-L & 300 & 0.69h & 3.0 & N/A & N/A & 6515.6 & N/A & 45.6 \\
LVBench & 103 & 1.14h & 15.0 & N/A & N/A & N/A & N/A & 45.4 \\
\hline
\textbf{\benchname}-M & 3,699 & 0.13h & 5.4 & 14,594.7 & 126.3 & 829.0 & 19,098.5 & 80.5 \\
\textbf{\benchname}-L & 649 & 0.72h & 30.8 & 83,183.1 & 720.0 & 4,724.7 & 109,286.1  & 80.5 \\
\textbf{\benchname}-XL & 50 & 9.32h & 400.0 & 1.079,716.9 & 9,345.5 & 61,326.7 & 1,418,995.0 & 80.5 \\
\textbf{\benchname}-UL & 12 & 38.85h & 1666.7 & 4,498,820.5 & 38,939.8 & 255,528.1 & 5,913,022.8 & 80.5 \\
\hline
\end{tabular}}
\caption{Benchmark statistics. As the level goes up (M/L/XL/UL: \tmedium, \tlong, \textralong, \tultralong), each video contains more QAs.
\textbf{\#Faces}: the number of facial bounding boxes per video.
\textbf{\#Spkrs}: the number of speaker labels (\textit{i.e.}, subtitle lines) per video.
\textbf{Sub/Cap/QA Length}: the tokens of all subtitles and captions per video, and the average tokens per question (options included).}
\label{tab:benchmark_statistics}
\end{table}

\clearpage

\section{QA Generation and Validation: Details}
\label{qa_generation}

\subsection{Initial QA Generation}
\label{qa_generation:initial_generation}

First, we define 28 detailed question categories. These 28 subcategories are then divided into three major groups for generation: low-level, high-level, and multi-point. For each subcategory, we provide a clear category definition and corresponding question examples. We show the definition and examples of a subcategory in Figure~\ref{fig:qa_type_define_examples}.
All category definitions and examples are summarized in the JSON file \texttt{qa\_type\_and\_samples.json}.

\begin{figure}[!h]
\centering
\fbox{
\begin{minipage}{0.9\linewidth}
\textbf{Body Movement} \\

\promptsection{Definition}
    \begin{enumerate}
        \item This task aims to assess the ability to capture the body movements and postures of individuals in the video. Therefore, questions in this dimension may involve asking about the type of actions performed by the person (such as standing, walking, waving, hugging), as well as the direction and magnitude of the movements (such as running to the left, raising arms high), etc.
    \end{enumerate}

    \promptsection{Example 1 }
    \begin{itemize}
        \item \textbf{Caption:} In the dance studio, the students are practicing fundamental skills, raising both hands above their heads in the `orchid fingers' position, slightly bending their knees in a half squat, tilting their bodies to the left, turning their heads toward the right shoulder, and directing their gaze toward the upper diagonal.
        
        \item \textbf{Question:} What are the trainees' physical movements like?
        
        \item \textbf{Options:} A. Arms hanging down, standing still B. Arms raised above the head with fingers in orchid position, half-squatting and leaning to the left C. Hands on hips, leaning to the right D. Arms crossed over chest, bending forward
        
        \item \textbf{Answer:} B
    \end{itemize}

    \promptsection{Example 2 }
    \begin{itemize}
        \item \textbf{Caption:} On the train platform, the mother waves goodbye to her son; she stands on tiptoe, raises her right hand high to wave a red scarf, covers her mouth with her left hand, leans slightly forward, and her gaze follows the slowly departing train.
        
        \item \textbf{Question:} How does the mother wave goodbye?
        
        \item \textbf{Options:} A. Arms hanging down, standing still
        B. Standing on tiptoe, raising and waving a red scarf, body leaning forward
        C. Hands on hips, shaking head
        D. Arms crossed over chest, turning away
        Answer: B. Stand on tiptoe, raise the red scarf high and wave it, body leaning forward.
        
        \item \textbf{Answer:} B
    \end{itemize}
    
\end{minipage}}
\caption{The definition and examples of  the \textit{Body Movement} subcategory.}
\label{fig:qa_type_define_examples}
\end{figure}

Based on the predefined question categories and corresponding examples, we generate question-answer pairs using the prompts shown in Figure~\ref{fig:qa_generation_prompts_part1} and Figure~\ref{fig:qa_generation_prompts_part2}. 

\begin{figure}[!h]
\centering
\fbox{
\begin{minipage}{0.9\linewidth}
\textbf{QA Generation Prompts, Part I} \\

    Given a description of a certain clip in a film or television drama, generate `Question-Option-Answer' triples based on the video description. These questions shall be applicable to the entire drama, and the understanding of video content shall be deepened through multi-dimensional Q\&A.
    The question stem shall not be too brief and must consist of two parts: the first part introduces the main information of the description, and the second part raises the question. 

    \promptsection{Guidelines for Generating Question - Option - Answer Triples}
    \begin{enumerate}
        \item \textbf{Core Requirements}:
        \begin{enumerate}
            \item Information Traceability: The answer shall clearly mark the corresponding key information basis in the video description (\textit{e.g.}, ``According to the plot of `after the XX incident' in the video……''), to ensure the traceability of reasoning basis.
            \item Content Fidelity: Strictly follow the core information such as scene settings, character relationships and event progression in the video description, and shall not add unmentioned details or inferences that deviate from the original content.
            \item Unambiguous Reference and Localization for Questioning: The question must precisely locate a unique scene in the video using time markers, location features, and event anchors. The referenced content should be clearly identifiable within the full video. Vague phrases such as `the current clip', `this content', `the beginning/opening/ending of the video', and `video description' are not allowed.
            \item Comprehensive Coverage: The generated triples shall evenly cover different stages, scenes and core events of the video description, avoiding concentration on a single plot to ensure the overall coverage of video content.
            \item Uniqueness of Answer: There is exactly one correct answer among the options, which is clear and objectively existing, and does not involve subjective evaluation or open-ended interpretation.
        \end{enumerate}
        \item \textbf{Option Design}:
        \begin{enumerate}
            \item Each question is set with 6-8 options to ensure a unique correct answer and a balanced distribution of options.
            \item Relevance: Distractors shall belong to the same logical category as the correct answer (\textit{e.g.}, homogeneous tools, similar scenes, related concepts)
            \item Discriminability: There shall be an essential difference from the correct answer to ensure it cannot be a reasonable answer.
            \item Rationality: Distractors themselves shall be reasonably possible options when separated from the video context.
            \item Misleading: Distractors shall have superficial credibility and can arouse reasonable doubt.
        \end{enumerate}
    \end{enumerate}

\end{minipage}}
\caption{The prompt to generate QA pairs (Part I).}
\label{fig:qa_generation_prompts_part1}
\end{figure}

\begin{figure}[!h]
\centering
\fbox{
\begin{minipage}{0.9\linewidth}
\textbf{QA Generation Prompts, Part II} \\

    \promptsection{Guidelines for Generating Question - Option - Answer Triples}
    \begin{enumerate}
        \setcounter{enumi}{2} 
        \item \textbf{Question Design}:
        \begin{enumerate}
            \item  Natural Expression of Questions: The expression shall be fluent, logically coherent, in line with human daily questioning habits, and without deliberate stiffness.
            \item The question stem must contain event reference and localization information, see the requirement of Unambiguous Reference and Localization for Questioning. It can include multiple elements such as time, location and events, and the reference information shall be slightly detailed.
            \item Ask questions directly targeting the video content, and do not ask questions about the `video description' itself (Do not use expressions such as ``Is XX mentioned in the description?'', ``According to the video description''; the correct form is ``Is XX in the video?'').
            \item The questions shall be suitable for questioning the entire drama, so do not generate questions inapplicable to the entire drama such as those targeting ``the beginning, ending, middle of the video''. Try to refer to the general content of the current clip in the question stem as much as possible.
            \item Avoid answer leakage in the question stem: Do not directly or indirectly imply or list the answer content in the question. Incorrect example: ``How many different scenes appear at the memorial service (Tiananmen Square, classroom, auditorium rostrum, auditorium side wall, auditorium entrance)?''
            \item Do not generate subjective and open-ended questions: Answers rely on subjective judgment, and different people may draw different conclusions, e.g., ``If the scenes described in the video are divided into two stages, what is the most reasonable dividing point?'', ``Which vehicle has the highest recognition due to its unique logo?''
            \item These triples shall cover each question dimension as much as possible, and each dimension can contain multiple `Question-Option-Answer' triples according to the amount of description content.

            \item Question Dimensions: The following are some question dimensions and reference samples for reference:\\
            \{task\_definitions\}\\
            \{demonstrations\}
        \end{enumerate}
    \end{enumerate}
    
    \promptsection{Output Format:}
    
        Your reply shall be as follows:
        [{{`Dimension': <Dim-1>, `Question': <q1>, `Options', <A.XXX B.XXX C.XXX D.XXX>, `Answer': <C>}}...]
\end{minipage}}
\caption{Continuing Figure~\ref{fig:qa_generation_prompts_part1}. The prompt to generate QA pairs (Part II).}
\label{fig:qa_generation_prompts_part2}
\end{figure}

\subsection{Rule and LLM-based Filtering}
\label{qa_generation:filtering}

\noindent \textbf{Rule filtering.} Through manual inspection of the generated QA pairs, we summarize a set of \textbf{filtering rules} to eliminate low-quality samples. The rules are as follows:
\begin{itemize}

    \item Remove QA pairs whose question contains any of the following phrases: \\
    \textit{the beginning of the video}, \textit{the start of the video}, \textit{the opening of the video}, \textit{the end of the video}, \textit{the last part of the video}, \textit{in the scene}, \textit{the entire video}, \textit{the overall narrative}, \textit{the first half of the video}, \textit{the second half of the video}, \textit{the start of the scene}, \textit{the end of the scene}

    \item Remove QA pairs whose question contains any of the following terms: \\
    \textit{subtitle}, \textit{Subtitle}, \textit{what is the function}, \textit{what function}

    \item Remove QA pairs that satisfy: \\
    Question category $\in \{\textit{counting}, \textit{quantity comparison}\}$ AND the question contains \textit{explicitly}, \textit{dialogue}, or \textit{mention}

    \item Remove QA pairs whose question contains any of the following phrases: \\
    \textit{video description}, \textit{explicitly mentioned}, \textit{explicitly stated}, \textit{described as}, \textit{entire description}

    \item Remove QA pairs if:
    \begin{itemize}
        \item The answer contains \textit{not mentioned}, \textit{not described}, \textit{not specified}, \textit{not indicated}; OR
        \item The question category is \textit{counting} and the answer contains words like \textit{several times} or \textit{some people}.
    \end{itemize}

    \item Remove QA pairs for which a large language model (LLM) determines a high probability of multiple correct answers.
\end{itemize}

\noindent \textbf{LLM-based filtering.}
To comprehensively evaluate the multi-modal abilities of models and prevent answers from being trivially deduced from the questions and options, we tested the direct response results of multiple language models. The prompts employed in the test are in Figure~\ref{fig:llm_eval_prompts}.

\begin{figure}[!h]
\centering
\fbox{
\begin{minipage}{0.9\linewidth}
\textbf{Prompts of LLM-based filtering} \\
\textbf{Task:} Please answer the following multiple-choice question.\\
\textbf{Instructions:}
\begin{enumerate}
    \item The question provides several options, but only one is correct. Please identify the correct option.
    \item  Note: Only return the letter of the correct answer (\textit{e.g.}, A, B, C, D, E, F, G, H). Do NOT include any reasoning or explanation.
\end{enumerate}
\end{minipage}}
\caption{The prompt of LLM-based filtering.}
\label{fig:llm_eval_prompts}
\end{figure}

\subsection{Human Checking and Editing}
\label{qa_generation:human_annotation}

\noindent \textbf{Annotation guidelines.}
Given a video of L2 sgements and multiple single-choice `Question-Option-Answer' triples related to the video, to help annotators understand the video content, the provided video clip is marked with face boxes (red boxes + names) for key characters. Annotators are required to check and determine whether each question fully meets the requirements and can be `retained' based on the video content. For data that does not meet the requirements, annotators should edit and correct it as much as possible to make it eligible for retention; for questions that are indeed unsuitable for correction, they should be marked as `not retained'.

\noindent \textbf{Retention criteria.}
Those QA pairs are retained only if they meet three criteria: the question accurately describes the video content (events, actions, details, characters), exactly one correct option exists, and the provided answer is correct.

\noindent \textbf{Editing criteria.}
The basic annotation principle is to edit and correct invalid triples as much as possible for retention, with detailed editing rules as follows: (1) Correct the answer if the question and options are valid but the answer is wrong; (2) Remove redundant options to retain only one correct one and confirm the answer if the question is valid but multiple correct options exist; (3) Revise or add the correct option if the question is valid but no correct option exists while a clear answer is available in the video; (4) Revise the question description and ensure answer correctness if the question is invalid but options are reasonable. Triples are marked as `not retained' if: the question is invalid and options are invalid (no/multiple correct answers); the answer cannot be obtained from the video; duplicate questions (different wording, identical intent) appear under the same video (starting from the second duplicate); the question is ambiguous, ungrammatical, hard to revise, or requires modifying over 50\% of the original text.

\subsection{Language Prior Removal}
\label{qa_generation:language_prior_removal}

For manually returned results, to further reduce the impact from language model priors and increase the difficulty of final evaluation results, we call language models again for evaluation without reference information. The prompt used is consistent with that in the second step `Rule and LLM-based Filtering', as shown in Figure~\ref{fig:llm_eval_prompts}. Three powerful language models are adopted here: Gemini-2.5-pro, GPT-5, and Seed-1.8, and cases where all three models make wrong predictions are retained. Some filtered examples are shown in Figure~\ref{fig:language_prior_removal_examples}.

\begin{figure}[!h]
\centering
\fbox{
\begin{minipage}{0.9\linewidth}
\begin{itemize}
    \item \textbf{Question 1:} During the conversation in the servants' hall, what items of clothing specifically identify Thomas and Jimmy as footmen in formal livery?
    \item \textbf{Options 1:}
        A. White aprons and black boots
        B. Grey trousers and white shirts
        C. Black tailcoats and white bow ties
        D. Brown vests and tweed caps
        E. Blue jackets and matching hats
        F. Green breeches and buckled shoes
    \item \textbf{Answer 1:} C

    \item \textbf{Question 2:} When Charles Grigg arrives at Isobel Crawley's house, from which part of the vintage Ford taxi does he exit?
    
    \item \textbf{Options 2:}
        A. The back seat, driver's side
        B. The back seat, passenger's side
        C. The front seat
        D. The driver's seat
        E. The luggage rack
        F. The running board
        
    \item \textbf{Answer 2:} B
\end{itemize}
\end{minipage}}
\caption{The examples of language prior removal.}
\label{fig:language_prior_removal_examples}
\end{figure}

\begin{figure}[!th]
\centering
\fbox{
\begin{minipage}{0.9\linewidth}
\textbf{Prompts of Fuzzy Matching} \\
Task:\\
Given a set of keywords A and a description B, determine whether description B contains the content corresponding to all keywords in A.\\

Output `Yes' if all keywords are included; if any keyword is missing, output `No' followed by the missing keywords. (No justification is required.)\\

Note: If description B contains synonyms or semantically similar expressions of the keywords in A, they shall be regarded as containing the corresponding information. For example:
\begin{enumerate}
    \item If the keyword is `kitchen', and B includes `cooking' or `cutting vegetables';
    \item If the keyword is `Zhou Bingkun', and B includes `Bingkun', `Kun'er', or `Director Kun';
    \item If the keyword is `kitchen', and B includes `cooking' or `cutting vegetables'.
\end{enumerate}
\end{minipage}}
\caption{The prompt of fuzzy matching.}
\label{fig:fuzzy_match}
\end{figure}


\subsection{Global Uniqueness Check}
\label{qa_generation:global_uniqueness}

To ensure the selected questions can be uniquely localized within the corresponding drama, we propose a multi-tool collaborative verification pipeline, which consists of the following three steps:

\noindent \textbf{Step 1: candidate segment construction via dual text retrieval.}
We first adopt two independent retrieval strategies to obtain L2 segments highly relevant to the target question, and merge them into a unified candidate set:
\begin{itemize}
    \item \textbf{BM25 retrieval~\cite{robertson2009probabilistic}}: For each drama, we build a retrieval index based on the captions of all L2 segments. We compute the relevance score between each question and the indexed caption texts, and retain the top 25 most relevant L2 segments.
    \item \textbf{Embedding similarity retrieval}: We use the pre-trained language model bge-large-en/zh-v1.5~\cite{xiao2024c} to generate embedding vectors for all L2 segment captions, where the chunk size in the text splitter is set to 200. We then calculate the cosine similarity between the question embedding and all caption embeddings, and select the top 25 most similar L2 segments.
\end{itemize}
The two retrieved segment sets are merged. After excluding the segment where the question is originally annotated, we obtain a candidate set containing up to 50 segments.

\noindent \textbf{Step 2: keyword extraction.}
We employ a large language model (LLM) to extract 3–5 key works from each question, which serve as the core semantic cues for uniqueness verification.

\noindent \textbf{Step 3: fuzzy matching.}
We use an LLM (Qwen-3-32B) to examine the caption of each candidate L2 segment. The prompt used here is showed in Figure~\ref{fig:fuzzy_match}. If none of the candidate captions contain all the extracted key words, the question is regarded as uniquely locatable in the entire drama. Otherwise, its localization uniqueness is not confirmed.

\clearpage

\section{Experimental Results}
\label{experimental_results}

\begin{figure}[!t]
\centering
\includegraphics[width=\linewidth]{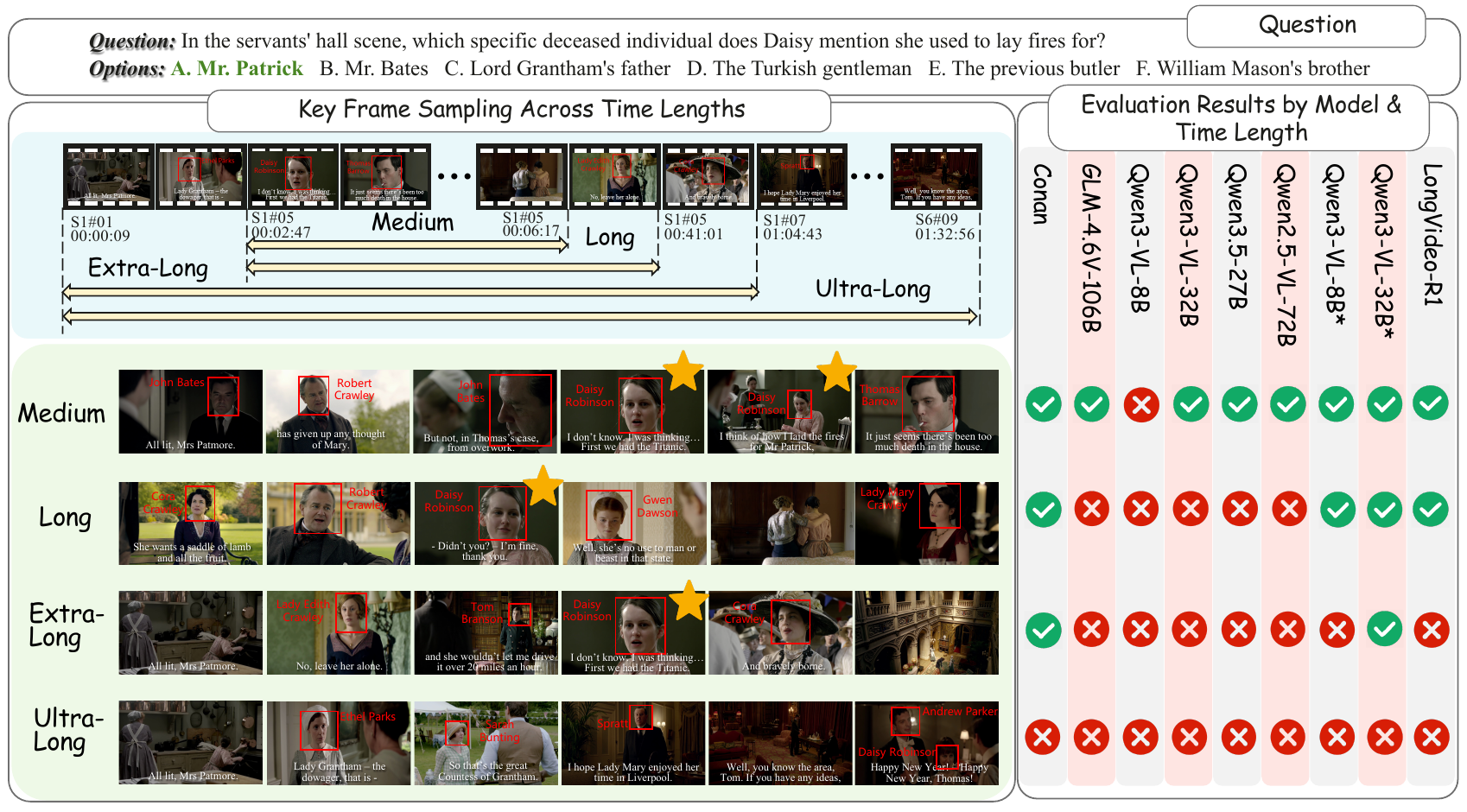}
\caption{Case study of an example. As the input video duration increases, the hit ratio (detecting a relevant frame, marked with \textcolor[HTML]{B8860B}{yellow stars}) of random sampling decreases significantly; consequently, an increasing portion of MLLMs produce incorrect answers. $^\star$: indicating that model thinking is enabled.}
\label{fig:failure_cases}
\end{figure}

\subsection{Setting}
\label{experimental_results:setting}

\noindent \textbf{Uniform sampling.}
For each TV drama, we first calculate the ID for all frames. The starting and ending parts of each episode are removed beforehand. Therefore, each input video, no matter how long it is, is defined by a start and end frame ID. This makes it easy to sample frames evenly without needing to join all episodes into one single video file (note that L3 to L6 units can span multiple episodes).

\noindent \textbf{Video input preparation.}
Each sampled frame is rendered with both the text and facial IDs. To avoid repeating information, please see Appendix~\ref{data_preparation:captioning} for details.

\noindent \textbf{Hierarchical captioning for agentic MLLMs.}
This part is very similar to Appendix~\ref{data_preparation:captioning}, but there are two differences. (1) The text transcript and speaker identities are \textbf{not} included in the prompt (only the rendered text on the frame is available); this ensures a fair comparison with the native MLLMs. (2) Following~\cite{qiu2026longvideo}, we define the time units from top to bottom (instead of the bottom to top method used earlier). This is because the earlier method assumes the base captions are already extracted, which is not true for this test.

In practice, starting from the main video, we evenly divide each part into $K$ smaller parts, and repeat this for 3 levels. This means there are $K^3$ parts at the lowest level. To use the pre-trained model from~\cite{qiu2026longvideo}, we set $K=\min\{6,\sqrt[3]{T/16\mathrm{s}}\}$, where $T$ is the length of the video in seconds. We generate captions by sampling $64$, $32$, and $16$ frames evenly for the clips in the first (highest), second, and third (lowest) levels.

\noindent \textbf{Configuration of Native MLLMs.}
We run all these models on a server with 8 NVIDIA A800 GPUs using the vLLM library. All models use the default settings and the same QA prompt, as shown in Figure~\ref{fig:llm_eval_prompts}. This QA prompt is the same one used for filtering easy questions, except (1) we add video tokens as context, and (2) we turn on the thinking feature for each specific model. During the test, we use different thread numbers for different models to reach the highest speed. However, to measure the time cost fairly, we set the thread number to 1 and report the average time across 100 samples.

\subsection{\textcolor{red}{Detailed Results}}
\label{experimental_results:detailed_results}

\begin{table}[!t]
\centering
\caption{Mapping of Major Categories to Subcategories and Statistics of QA Numbers.}
\label{tab:category_subcategory_mapping}
\resizebox{\linewidth}{!}{
\begin{tabular}{l|c|l|c}
\toprule
\textbf{Major Category} & \textbf{QA Number (Full/Mini)} &
\textbf{Subcategory} & \textbf{QA Number (Full)} \\
\midrule
\multirow{8}{*}{Visual Perception} & \multirow{8}{*}{4966 / 475}
& Character Clothing (CCI) & 1397 \\
& & Character Hairstyle (CH) & 306 \\
& & Distinctive Features (DF) & 399 \\
& & Object Appearance (OA) & 1168 \\
& & Object Presence (OP) & 437 \\
& & Season and Weather (SAW) & 155 \\
& & Scene and Location (SAL) & 498 \\
& & Textual Content (TCo) & 606 \\
\midrule
\multirow{3}{*}{Spatial Understanding} & \multirow{3}{*}{1409 / 147}
& Spatial Location (SL) & 704 \\
& & Position Comparison (PC) & 620 \\
& & Shooting Perspective (SP) & 85 \\
\midrule
\multirow{4}{*}{Action and Interaction} & \multirow{4}{*}{5290 / 553}
& Body Movement (BM) & 1044 \\
& & Action Claim Subject (ACS) & 2852 \\
& & Action Recipient (AR) & 1219 \\
& & Object Motion (OM) & 175 \\
\midrule
\multirow{4}{*}{Event Understanding} & \multirow{4}{*}{1981 / 195}
& Event Summary (ES) & 309 \\
& & Event Background (EB) & 929 \\
& & Event Outcome (EO) & 361 \\
& & Character Motivation (CM) & 382 \\
\midrule
\multirow{4}{*}{Temporal Understanding} & \multirow{4}{*}{2710 / 275}
& Character Change (CC) & 230 \\
& & State Transition (ST) & 348 \\
& & Temporal Sequence (TS) & 1019 \\
& & Temporal Comparison (TC) & 1113 \\
\midrule
\multirow{3}{*}{Relation \& Logical Reasoning} & \multirow{3}{*}{1868 / 185}
& Human-Object Relation (HOR) & 996 \\
& & Character Relation (CR) & 284 \\
& & Causal Inference (CI) & 588 \\
\midrule
\multirow{2}{*}{Counting} & \multirow{2}{*}{1776 / 170}
& Object Counting (OC) & 1325 \\
& & Quantity Comparison (QC) & 451 \\
\midrule
\textbf{Total} & \textbf{20000 / 2000}
& \textbf{All Subcategories} & \textbf{20000} \\
\bottomrule
\end{tabular}
}
\end{table}

\begin{figure}
    \centering
    \includegraphics[angle=-90,width=\linewidth]{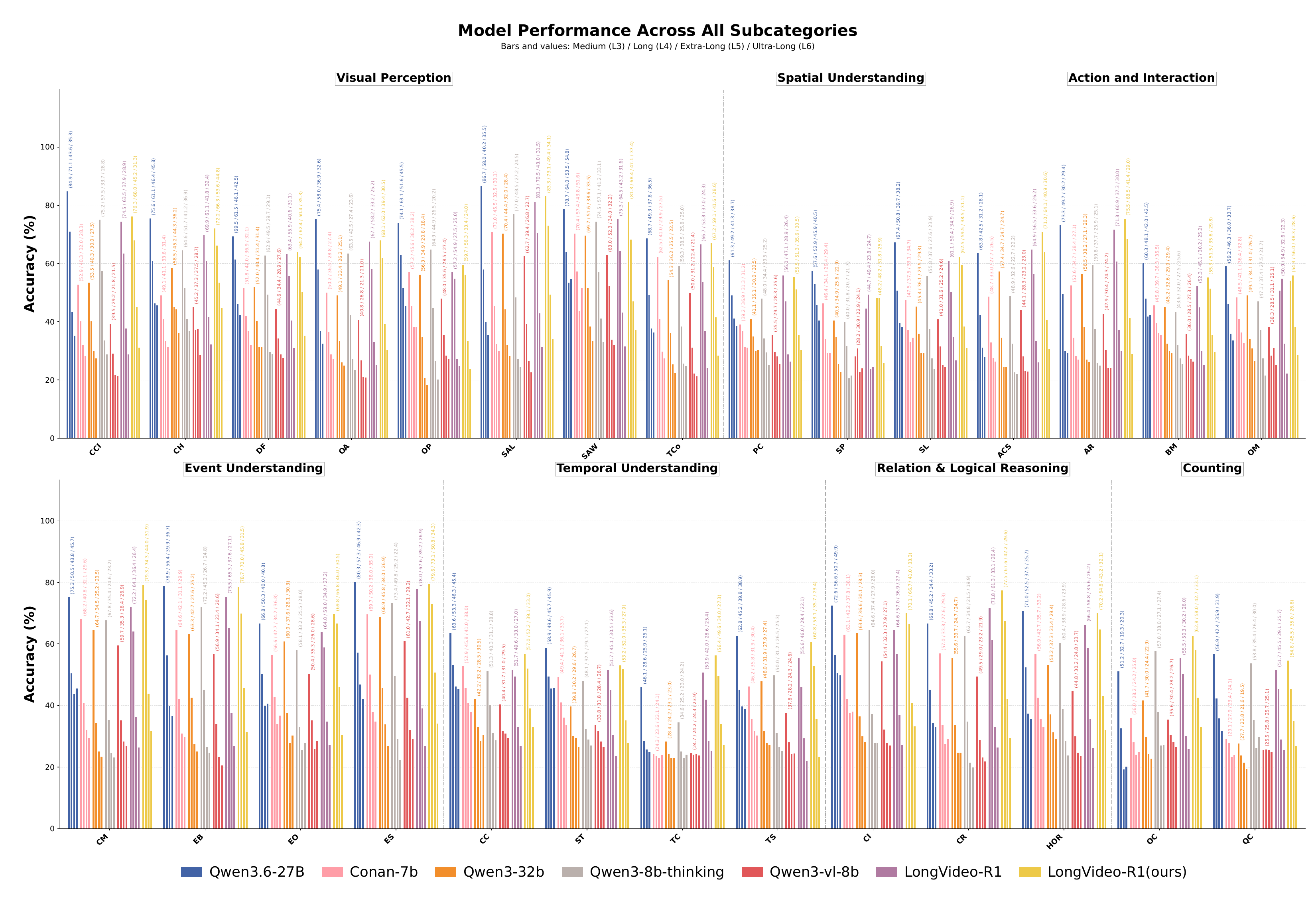}
    \caption{Model performance comparison across all sub-category dimensions. The frame number of the Native MLLMs is 256. (Best viewed zoomed in and in color.)}
    \label{fig:resultcategories256}
\end{figure}

\begin{figure}
    \centering
    \includegraphics[angle=-90,width=\linewidth]{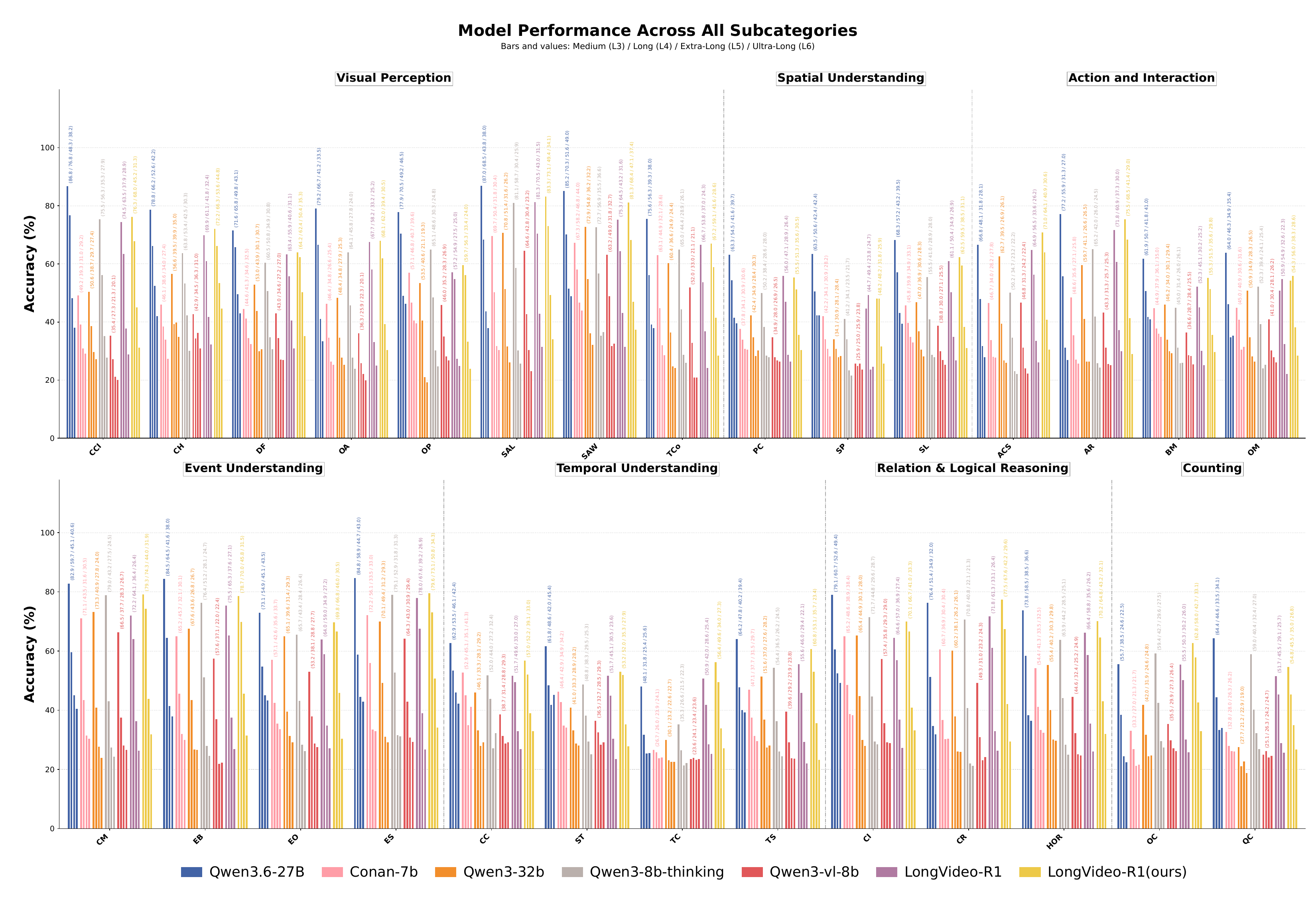}
    \caption{Model performance comparison across all sub-category dimensions. The frame number of the Native MLLMs is 512. (Best viewed zoomed in and in color.)}
    \label{fig:resultcategories512}
\end{figure}

\noindent \textbf{Category-wise.}
Figure~\ref{fig:resultcategories256} and Figure~\ref{fig:resultcategories512} show the performance comparison of different models on fine-grained subcategories over the entire Full dataset. In Figure~\ref{fig:resultcategories256}, native MLLMs (e.g., the Qwen series) are evaluated with 256 sampled frames, whereas in Figure~\ref{fig:resultcategories512}, they use 512 sampled frames. Each color corresponds to one model, and the color-to-model mapping is provided in the figure legend. The four adjacent histogram bars and their annotated performance values represent the results on \tmedium, \tlong, \textralong, and \tultralong sequences, respectively.
The fine-grained categories are shown in abbreviated form, and their full names are listed in Table~\ref{tab:category_subcategory_mapping}. This table also summarizes the hierarchical relationships between coarse-grained and fine-grained categories, as well as the number of QA pairs for each category and subcategory. For the seven major categories, we report the number of QA pairs on both the Full set and Mini set, allowing researchers to choose different evaluation sets according to their computational resources.

\noindent \textbf{Input-Condition Ablation Study.}

To investigate the effects of different prompting strategies and input conditions, we evaluate Qwen3-32B on Medium (L3) questions under different input
conditions, as shown in Table~\ref{tab:input_ablation}.
For subtitles, we compare hard-rendering them onto video frames with directly
providing the subtitle text. Providing subtitles as text slightly outperforms hard-rendering them onto video frames (51.0\% vs.\ 50.0\%). We therefore adopt direct subtitle text, which also aligns with the protocols of existing video benchmarks.
For character-role matching, we compare rendered face bounding boxes with a
character gallery containing one image per character. Rendering face bounding
boxes performs better, suggesting that it provides a more effective
representation of character identities.
Incorporating subtitle text yields a 12.4\% performance gain, demonstrating that subtitles are critical for video understanding, especially for narrative content. Speaker identity information further brings a 2.0\% improvement. Such annotations are rarely available in other video understanding datasets and constitute a key contribution of our benchmark.

\begin{table}[t]
    \centering
    \label{tab:input_ablation}
    \setlength{\tabcolsep}{5pt}
    \begin{tabular}{l c c c  c l}
        \toprule
        \textbf{Face} &
        \textbf{Subtitle} &
        \textbf{Speaker} &
        \textbf{Acc.} \\
        \midrule
         bounding boxes
            & hard-rendered
            & \xmark  & 50.0 \\
         bounding boxes
            & text
            & \xmark  & 51.0 \\
         character gallery
            & text
            & \xmark  & 49.4 \\
         bounding boxes
            & \xmark
            & \xmark  & 38.6 \\
         bounding boxes
            & text
            & \cmark  & \textbf{53.0} \\
        \bottomrule
    \end{tabular}
    \caption{Performance under Different input conditions.The character gallery contains the first observed face image for
    each character.}
\end{table}

\subsection{Qualitative Studies}
\label{experimental_results:qualitative_studies}

Figure~\ref{fig:failure_cases} shows a typical test case. The experimental results in the text demonstrate that the reasoning model achieves higher accuracy when answering the same questions.
Below is a concrete example: we use Qwen3-VL-8B~\cite{bai2025qwen3} to answer the same question with and without the reasoning mode enabled.
It can be clearly observed that the response quality is significantly higher under the reasoning mode.
\begin{itemize}
    \item \textbf{Question:} After Rita leaves and the initial tense silence is broken, who does Robert identify as the person he is most disappointed in?
    \item \textbf{Options:} A. Himself  ~\hspace{2em} B. Rita Bevan 
        ~\hspace{2em} C. Mary 
    
        ~\hspace{2em} D. Tony Gillingham 
       ~\hspace{1em} E. His mother, Violet ~\hspace{1em} F. His wife, Cora
    \item \textbf{Answer:} D 
    \item \textbf{Response w/o reasoning:} [A]

    \textbf{Response w/ reasoning:} In the provided video clip, the key scene occurs after the conversation between Robert Crawley and Rita Bevan. According to the subtitles and visual content, Robert explicitly expresses his disappointment in Tony Gillingham during his subsequent conversation with Lady Mary Crawley. Specifically, in the clip, Robert Crawley says: ``I find I'm most disappointed in Tony Gillingham...'' This statement directly identifies the person he is most disappointed in, as asked in the question.
    
    Therefore, the correct answer is: [D]
    
\end{itemize}

\noindent When the model is unable to answer the question based on the provided information, it will randomly select an option. Below is a concrete example: we use Qwen-VL-8B~\cite{bai2025qwen3} to respond to a question. For demonstration purposes, only the answer under the reasoning mode is presented.
\begin{itemize}
    \item \textbf{Question:} In the bathroom, Molesley briefly considers using another product to treat his hair. What is the described color of the liquid in the bottle he looks at?
    
     \textbf{Options:} A. Pink  ~\hspace{1.2em} B. Blue
        
         ~\hspace{4em} C. Yellow  \hspace{0.7em} D. Clear
        
       ~\hspace{4em} E. White  ~\hspace{1em}F. Green

    \textbf{Answer:} C

    \textbf{Response w/ reasoning:} In the provided video frames, there is no scene relevant to the question. The question asks about the color of the liquid in the bottle that Molesley considers using for his hair while in the bathroom. However, the provided screenshots primarily depict various characters from Downton Abbey engaged in conversations and interactions in settings such as the kitchen, drawing room, bedroom, and hallway. None of these images show Molesley in a bathroom, nor do they depict him looking at or considering any hair treatment product.

    Therefore, based on the provided screenshots, it is impossible to answer this question. I will randomly select an answer: [D]
    
\end{itemize}

\clearpage

\section{Details of Accuracy-Efficiency Tradeoff}
\label{tradeoff}

This section compares native MLLMs and agentic algorithms regarding the accuracy-efficiency trade-off. Recall that the Naive-RAG methods and LongVideo-R1 in Table~\ref{tab:main_results} use hierarchical video captions generated by Qwen3-VL-32B. Therefore, we mainly compare them against the direct QA results of that same model. Across all tracks, the Naive-RAG algorithms show a clear advantage over native MLLMs, but they take much longer to run (\textit{e.g.}, 293.0 seconds per \tmedium video, and 1,664.6 seconds per \tultralong video, which is 1--2 orders of magnitude longer than the execution time of native MLLMs). This is because they process more sampled frames and make more video captioning calls. Two methods can reduce this computational burden: using beam search to narrow the search scope, or adding a reasoning LLM (\textit{e.g.}, LongVideo-R1) for smart navigation. Both methods greatly reduce the time complexity, especially on the \tultralong track (by a factor of 6--8). LongVideo-R1 produces higher QA accuracy, but it adds noticeable costs for the LLM reasoning steps. We summarize the QA accuracy and inference times of these models in Table~\ref{tab:tradeoff}.

\begin{table}[!t]
\setlength{\tabcolsep}{0.12cm}
\centering
\resizebox{0.8\linewidth}{!}{
\begin{tabular}{|l|c|cc|cc|cc|cc|}
\hline
\multirow{2}*{\textbf{Model}} & \multirow{2}*{\textbf{TP}} & \multicolumn{2}{c|}{\tmedium} & \multicolumn{2}{c|}{\tlong} & \multicolumn{2}{c|}{\textralong} & \multicolumn{2}{c|}{\tultralong} \\
{} & {} & Acc. & \textit{time} & Acc. & \textit{time} & Acc. & \textit{time} & Acc. & \textit{time} \\
\hline
\rowcolor{lightgray} \multicolumn{10}{|l|}{\textit{Open-source Native MLLMs}} \\
Qwen3-VL-32B$^\dagger$$^\ddagger$ & C/B & 51.0 & \textit{22.6} & 35.8 & \textit{36.4} & 28.2 & \textit{37.6} & 34.1 & \textit{30.4} \\
\hline
\hline
\rowcolor{lightgray} \multicolumn{10}{|l|}{\textit{Open-source Agentic MLLMs}} \\
\multirow{2}*{Naive-RAG (Q3.8)} & C & \multirow{2}*{60.1} & \textit{293.0} & \multirow{2}*{56.7} & \textit{1,664.6} & \multirow{2}*{37.0} & \textit{1,955.5} & \multirow{2}*{31.5} & \textit{1,955.0} \\
{} & B & {} & \textit{34.9} & {} & \textit{34.1} & {} & \textit{4.9} & {} & \textit{1.4} \\
\hline
\multirow{2}*{RAG-BS (Q3.8)} & C & \multirow{2}*{56.4} & \textit{180.5} & \multirow{2}*{40.4} & \textit{372.2} & \multirow{2}*{32.1} & \textit{408.3} & \multirow{2}*{30.9} & \textit{406.5} \\
{} & B & {} & \textit{26.7} & {} & \textit{15.3} & {} & \textit{4.1} & {} & \textit{1.3} \\
\hline
\multirow{2}*{LongVideo-R1} & C & \multirow{2}*{62.8} & \textit{105.3} & \multirow{2}*{54.8} & \textit{169.1} & \multirow{2}*{34.7} & \textit{294.1} & \multirow{2}*{27.2} & \textit{322.1} \\
{} & B & {} & \textit{36.1} & {} & \textit{41.2} & {} & \textit{41.8} & {} & \textit{33.5} \\
\hline
\end{tabular}
}
\caption{Full-set QA accuracy (\%) and inference time (in seconds) with different testing protocols (TPs), case-wise (C) or batch-wise (B). Note that case-wise numbers are identical to those in Table~\ref{tab:main_results}. 
$^\dagger$: with 256 frames;
$^\ddagger$: model thinking is disabled.
}
\label{tab:tradeoff}
\end{table}

Furthermore, unlike other long-video benchmarks, \benchname contains many questions for each input video, especially in the \textralong and \tultralong tracks. When processing multiple questions together, agentic algorithms (including RAG) can share the computation cost of video captioning across many queries. As shown in Table~\ref{tab:tradeoff}, this shifts the accuracy-efficiency trade-off. Naive-RAG and RAG-BS become as efficient as native MLLMs on the \tmedium and \tlong tracks, and even an order of magnitude faster on the \textralong and \tultralong tracks. Meanwhile, LongVideo-R1 reports comparable time costs while achieving higher QA accuracy. Overall, agentic algorithms become the better choice for batch-wise tasks, even though they require a heavy pre-processing stage to generate the hierarchical captions.

This analysis opens up several new questions: (1) Is there a better data structure than hierarchical video captions to further improve this trade-off? (2) How can we adapt agentic algorithms to achieve better performance for batch, streaming, and incremental QA tasks? (3) Can these learned abilities transfer across different TV dramas and to new video categories, such as egocentric videos? Studying these problems connects long-form video understanding with agentic memory mechanisms~\cite{yao2025survey,hu2025memory,abou2025agentic}, enlightening new research directions.

\noindent \textbf{Time cost estimation.} Below we explain how we estimate the time cost of different models in Table~\ref{tab:tradeoff}.
\begin{itemize}
\item \textbf{The native MLLMs}, \textit{e.g.}, Qwen3-VL-32B. We use the same estimation method stated before (setting the thread number to 1 and averaging over 100 samples). Native MLLMs cannot share computation among multiple test cases. We cannot feed multiple QAs into the native MLLMs at once (with the same video unit as context) because this may cause answer leaking: the answer to some QAs may be included in the description or options of other QAs.
\item \textbf{The agentic MLLMs}, \textit{i.e.}, Naive-RAG, RAG-BS, and LongVideo-R1. They share the same process of hierarchical captioning for each input video unit. Table~\ref{tab:main_results} reports the time cost as if each QA is executed individually, and Table~\ref{tab:tradeoff} reports the amortized cost where the captioning cost is averaged across all QAs in the input video unit. For the calculation, we estimate that the time used for Qwen3-VL-32B video captioning of the 1st, 2nd, and 3rd-level clips (with 64, 32, and 16 frames, respectively) is $14.18$s, $11.16$s, and $6.82$s. These are tested on a server of 8 NVIDIA A100 GPUs. The time cost for building an inverted table and performing BM25 search is about $0.3$-$0.6$s for each test case. LongVideo-R1 takes 30 to 40 seconds to navigate through the hierarchical caption (by calling the $\mathtt{video\_cap}$ tool) and performs QA (by calling the $\mathtt{video\_qa}$ tool) if necessary.
\end{itemize}

\clearpage

\section{Details of Multi-Hop Composition}
\label{riddles}


\begin{figure}[!t]
\centering
\includegraphics[width=0.95\linewidth]{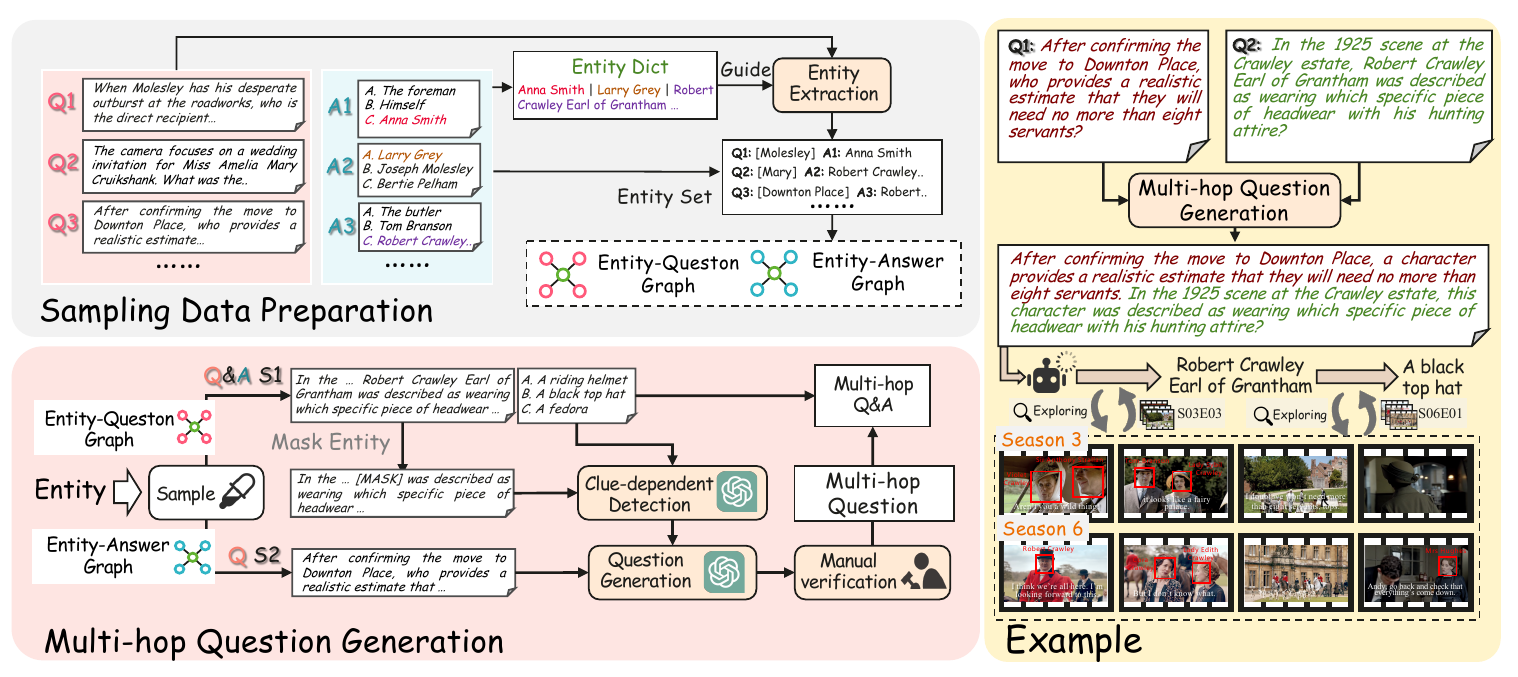}
\vspace{-15pt}
\caption{Illustration of constructing a two-hop riddle from the basic QAs. In this example, the two building blocks come from two different seasons. Hence, the composed QA naturally requires ultra-long context understanding.}
\vspace{-5pt}
\label{fig:riddle_examples}
\end{figure}

\begin{figure}[!h]
\centering
\fbox{
\begin{minipage}{0.9\linewidth}
\textbf{Prompt for Multi-hop Question Validity Verification} \\
You are an expert in multi-hop question validity verification.

Task: Determine whether the bridge entity \texttt{[MASK]} is necessary for answering the question, and whether the options allow the identity of \texttt{[MASK]} to be easily inferred.


\textbf{Evaluation Criterion 1: Highly strict invalid cases:}
\begin{enumerate}
\item \texttt{options\_leak}.  
The identity of \texttt{[MASK]} can be easily inferred solely from the options, or the candidate space is drastically narrowed so that \texttt{[MASK]} or the correct answer is almost determined.

\item \texttt{other\_clues\_sufficient}.  
If the question can still be uniquely and reliably answered after removing \texttt{[MASK]}, the question should be classified as invalid.
\end{enumerate}

\textbf{Evaluation Criterion 2: Conditionally invalid cases (only with very strong evidence):}
\begin{enumerate}
\item \texttt{scene\_keywords\_locate}.  
A combination of a character/title with explicit location or scene keywords that is sufficient to almost uniquely locate the event.

\item \texttt{dialogue\_quote\_plot\_locate}.  
Highly recognizable dialogue, direct quotes, or distinctive plot descriptions that directly identify the event or strongly imply the identity of \texttt{[MASK]}.

\item \texttt{context\_document\_infer}.  
Document-like contextual clues (e.g., reports, accusation letters, notices, number of bribes) that allow stable inference of the referenced entity (\texttt{[MASK]}) and the corresponding event.

\item \texttt{unique\_action\_phrase\_locate}.  
Very rare actions, terminology, or expressions that allow the event to be identified with high probability even without knowing \texttt{[MASK]}.
\end{enumerate}

\textbf{Output Format:}

Return the result strictly in the following JSON format:

\{
\begin{itemize}
\item[] ``\texttt{is\_reasonable}'': true or false
\item[] ``\texttt{invalid\_type}'': one of 
      \texttt{options\_leak},
      \texttt{other\_clues\_sufficient},
      \texttt{scene\_keywords\_locate},
      \texttt{dialogue\_quote\_plot\_locate},
      \texttt{context\_document\_infer},
      \texttt{unique\_action\_phrase\_locate},
      or \texttt{null}
\item[] ``\texttt{reason}'': A one-sentence explanation stating whether the question is reasonable or which invalid category applies and why. 
\end{itemize}
\}

\textbf{Input:}

Original Question: \{original\_question\}  

Masked Question: \{masked\_question\}  

Bridge Entity: \{bridge\_str\}  

\{answer\_line\}  

\{options\_line\}
\end{minipage}}
\caption{Prompt used for multi-hop question validity verification. 
The model determines whether the bridge entity \texttt{[MASK]} is necessary for answering the question and whether the options leak its identity.}
\label{fig:riddle_verification}
\end{figure}

\begin{figure}[!h]
\centering
\fbox{
\begin{minipage}{0.9\linewidth}
\textbf{Prompt for Two-hop Riddle Generation: Part I} \\

Please generate a two-hop reasoning question based on the following two related QA pairs, and return the result strictly in JSON format.

\textbf{Generation Goal:}

Construct a two-hop question that can only be answered by first resolving the prerequisite reasoning step (QA2). The question is designed for the full-series search space of a long-form TV drama (approximately 30 hours) and must satisfy both unique identifiability and a chained reasoning structure. The wording should be as natural as possible, avoiding awkward phrasing caused by excessively long stacked modifiers. 

\textbf{Requirements:}

\begin{enumerate}
    \item The new question must first use QA2 to identify a specific person or entity (the bridge entity, \{bridge\_ent\}), and then use that person or entity to answer the question corresponding to QA1.
    
    \item The new question must not explicitly mention ``\{bridge\_ent\}''. Instead, it must uniquely refer to this person or entity through a complete and specific description of the key scene, key utterance, key behavior, or key plot event in QA2.
    
    \item Do not semantically compress or generalize any critical information from QA2 or QA1:
    \begin{enumerate}
        \item Do not replace original key utterances with vague summaries.
        \item Do not remove plot details that contribute to unique identification.
        \item You must preserve the key scene, key action, key dialogue, key conflict, or key emotional state from the original questions.
        \item Reordering is allowed, but the information density must not be reduced and the meaning must not be blurred.
    \end{enumerate}
    
    \item The new question must not introduce any temporal order that is not explicitly given in the original QA pairs:
    \begin{enumerate}
        \item Do not use temporal markers such as ``later,'' ``afterward,'' ``then,'' ``subsequently,'' or similar expressions.
        \item If the temporal order is uncertain, use temporally neutral connections such as ``this person'' or ``this character \ldots when \ldots''.
    \end{enumerate}
    
    \item Do not force the entire QA2 content into a single oversized premodifier phrase (e.g., avoid structures like ``the person who \ldots, when \ldots, \ldots''). Instead, the question must adopt one of the following more natural forms:
    \begin{enumerate}
        \item \textbf{Narrative form:} first describe the QA2 event in a complete sentence, then use ``this person'' or ``this character'' to introduce the QA1 query.
        \item \textbf{Multi-clause form:} connect the content using two or three shorter clauses or sentences, rather than one overly long sentence.
    \end{enumerate}
    
    \item The new question must form a strict chained reasoning structure:
    \begin{enumerate}
        \item Without the QA2-related information, the new question must not be uniquely locatable across the full series.
        \item QA2 alone must not directly answer the final question.
        \item QA1 alone must not allow the answer to be obtained without going through QA2.
    \end{enumerate}
\end{enumerate}
\end{minipage}}
\caption{The prompt to generate two-hop riddles.}
\label{fig:riddle_generation_part1}
\end{figure}

\begin{figure}[!h]
\centering
\fbox{
\begin{minipage}{0.9\linewidth}
\textbf{Prompt for two-hop riddle Generation: Part II} \\

\begin{enumerate}
    \setcounter{enumi}{6} 
    \item The wording of the new question must be natural and fluent, consistent with how real humans ask questions:
    \begin{enumerate}
        \item Diverse sentence structures are encouraged, such as rhetorical questions, inversion, or postponed subjects.
        \item Avoid template-based concatenation and mechanical repetition.
        \item Priority should be given to logical rigor, semantic clarity, and information completeness.
    \end{enumerate}
    
    \item The new question must retain at least one key clause from the original QA2 question to ensure that the information anchor is preserved.
    
    \item The answer to the new question must be exactly the same as the answer to QA1.
    
    \item The output must be in JSON format and contain only one key: \texttt{``generated\_question''}.
    
    \item Return only the JSON string. Do not add any explanations, comments, or markdown formatting.
\end{enumerate}

\textbf{Example Output Format:}
\begin{quote}
\texttt{\{``generated\_question'': ``Example question text''\}}
\end{quote}

\textbf{Input:}

\textbf{Prerequisite Reasoning QA (QA2)}\\
Question: {qa2.question}\\
Answer: {qa2.answer\_text}

\vspace{0.5em}

\textbf{Final Reasoning QA (QA1)}\\
Question: {qa1.question}\\
Answer: {qa1.answer\_text}

\end{minipage}}
\caption{Continuing Figure~\ref{fig:riddle_generation_part1}. The prompt to generate two-hop riddles.}
\label{fig:riddle_generation_part2}
\end{figure}

\noindent \textbf{Overall objective.}
To make the problem more challenging, especially in evaluating the model's ability to answer questions by capturing long-range clues, we generate riddles that require multi-step reasoning over multiple pieces of evidence, based on basic QA pairs. Given the vast amount of information conveyed by characters in TV shows, we leverage character entities as bridging elements in this preliminary exploration. As shown in the right diagram of Figure~\ref{fig:riddle_examples}, the character Robert Crawley, Earl of Grantham, is the answer to question 1 ($\mathrm{Q}_1$) and also appears in the premise of question 2 ($\mathrm{Q}_2$). We remove Robert Crawley, Earl of Grantham, from $\mathrm{Q}_2$ and introduce this entity through $\mathrm{Q}_1$, thereby constructing a 2-hop question. To answer such questions, the model must first decompose the problem, answering the first sub-question by finding clues (from Season 3, Episode 3 of Downton Abbey) to obtain Robert Crawley, Earl of Grantham as an intermediate result, and then answer the second question (from Season 6, Episode 3) by finding clues, completing the multi-hop question.

\noindent \textbf{The detailed pipeline.}
The entire question construction process is illustrated in the left diagram of Figure~\ref{fig:riddle_examples}, consisting of two steps: Sampling Data Preparation and Multi-hop Question Generation. The purpose of Sampling Data Preparation is to establish the relationship graph between the bridging entities, questions, and answers. We first compile the correct answers (characters) from all basic QA pairs into an entity dictionary. Then, we iterate over all the questions in the basic QAs and establish the correspondence between the entities in the entity dictionary and these questions. This results in two data graphs: the Entity-Question Graph and the Entity-Answer Graph. In the Multi-hop Question Generation step, we randomly sample an entity and use these two graphs to find the corresponding QA pairs. Sampling from the Entity-Answer Graph yields an answer whose associated question forms the first question in the reasoning chain, while the question sampled from the Entity-Question Graph serves as the second question in the chain, with its corresponding answer being the final answer. However, the second question may contain some information leakage, where the model does not need to solve the first sub-question to obtain the bridging entity and can directly answer by only looking at the second sub-question (for example, ``Who is the son of \texttt{[MASK]} and B.'' In a TV show, the context is so limited that there is no need to know who \texttt{[MASK]} is to answer the question). To address this, we attempt to mask the bridging entity \texttt{[MASK]} and use LLMs to assess whether the question relies on the bridging entity to answer. If it doesn't, it indicates that there is redundant information leakage, and we resample the question. This process is guided by the prompt shown in Figure~\ref{fig:riddle_verification} to instruct GPT-5. Afterward, we use the prompt in Figure~\ref{fig:qa_generation_prompts_part1} and Figure~\ref{fig:qa_generation_prompts_part2} to combine the two questions into a two-hop question.

\noindent \textbf{Results.}
Using this pipeline, we generate 452 questions from the basic QA pairs for a preliminary evaluation. For evaluation, we use Qwen3-VL-32B with Medium track clips (approximately 20 minutes) provided as auxiliary context. The model achieves an accuracy of 24.8\%, which increases to 27.8\% when only the second-hop question is evaluated. This indicates that current models still struggle with multi-step reasoning over temporally distant clues. In future work, we plan to generate more challenging questions by leveraging richer video information, including annotations and other structured signals, enabling benchmarks that require integrating multiple clues across long temporal spans.



\end{document}